%% file: main.tex
\documentclass[review,3p,times,authoryear,11pt]{elsarticle}

\usepackage{tikz}

\usepackage{graphicx}
\usepackage{amssymb}
\usepackage{verbatim}
\usepackage{adjustbox}
\usepackage{algorithm}
\usepackage{algorithmic}
\usepackage{booktabs}
\usepackage{colortbl}
\usepackage{arydshln}
\usepackage{multirow}
\usepackage{rotating}
\usepackage{hyperref}
\usepackage{subfigure}
\usepackage{color}
\usepackage{clrscode3e}
\usepackage{longtable}
\usepackage{setspace}
\usepackage{threeparttable}
\usepackage[T1]{fontenc}	
\usepackage{chngcntr}
\counterwithout{figure}{section}
\usepackage{amsthm}
\usepackage{amsmath}

\usepackage{overpic}
\usepackage{amsfonts}

\newlength{\commentindent}
\usepackage{float}
\usepackage{diagbox}
\usepackage{lscape}
\usepackage{xcolor}
\usepackage{geometry}
\journal{}
 
\begin{document}

\begin{frontmatter}


\title{Large Neighborhood Search and Bitmask Dynamic Programming for Wireless Mobile Charging Electric Vehicle Routing Problems in Medical Transportation}
\author{Jingyi Zhao$^{1}$}
\author{Haoxiang Yang$^{2}$}
\author{Yang Liu$^{3*}$}

\address{$^{1}$ Shenzhen Research Institute of Big Data, The Chinese University of Hong Kong (Shenzhen), 518172}

\address{$^2$ School of Data Science, The Chinese University of Hong Kong, Shenzhen (CUHK-Shenzhen), Shenzhen 518172, China}
\address{$^3$ Department of Civil and Environmental Engineering  \& Department of Industrial Systems Engineering and Management, National University of Singapore, Singapore}

\cortext[corl]{Corresponding Author; +Equal Contribution}

\begin{abstract}
\input{Sections/0abstract}
\end{abstract}

\begin{keyword}
Electric vehicle routing \sep Mobile charging \sep Healthcare \sep Dynamic programming  \sep Meta-heuristic

\end{keyword}

\end{frontmatter}

\newpage
\input{Sections/1Introduction}
\input{Sections/2RelatedWork}
\input{Sections/3ProblemStatement}
\input{Sections/4DP}
\input{Sections/5Algorithm}
\input{Sections/6Experiment}
\input{Sections/7Conclusion}

\section*{Acknowledgements}
This work was supported by 
\bibliographystyle{model5-names} 

\bibliography{citation}

\newpage
\input{Sections/8Appendice}

\end{document}

%% file: Sections/0abstract.tex
The transition to electric vehicles (EVs) is critical to achieving sustainable transportation, but challenges such as limited driving range and insufficient charging infrastructure have hindered the widespread adoption of EVs, especially in time-sensitive logistics such as medical transportation.
This paper presents a new model to break through this barrier by combining wireless mobile charging technology with optimization. We propose the Wireless Mobile Charging Electric Vehicle Routing Problem (WMC-EVRP), which enables Medical Transportation Electric Vehicles (MTEVs) to be charged while traveling via Mobile Charging Carts (MCTs). This eliminates the time wastage of stopping for charging and ensures uninterrupted operation of MTEVs for such time-sensitive transportation problems. However, in this problem, the decisions of these two types of heterogeneous vehicles are coupled with each other, which greatly increases the difficulty of vehicle routing optimizations.
To address this complex problem, we develop a mathematical model and a tailored meta-heuristic algorithm that combines Bit Mask Dynamic Programming (BDP) and Large Neighborhood Search (LNS). 
The BDP approach efficiently optimizes charging strategies, while the LNS framework utilizes custom operators to optimize the MTEV routes under capacity and synchronization constraints. Our approach outperforms traditional solvers in providing solutions for medium and large instances.
Using actual hospital locations in Singapore as data, we validated the practical applicability of the model through extensive experiments and provided important insights into minimizing costs and ensuring the timely delivery of healthcare services.

%% file: Sections/1Introduction.tex
\section{Introduction}
\label{Section:Introduction}
As the world transitions to clean energy, the widespread adoption of electric vehicles (EVs) has become a key method for reducing carbon emissions and mitigating air pollution. However, despite significant advances in EV technology in recent decades, the application of EVs is still limited by battery range and charging time.  
Currently, fixed charging stations are the most commonly used method for charging EVs, but this method requires vehicles to be detoured to where there are charging stations, greatly reducing the viability of EVs in time-sensitive logistical problems such as medical transportation.

A number of innovative solutions have emerged in recent years to address the charging anxiety of EVs. 
Mobile charging vehicles have gained popularity as a flexible charging option, offering on-demand services. This approach is particularly beneficial for vehicles that have run out of battery power and cannot easily access a fixed charging station, especially in remote areas or regions with insufficient charging infrastructure. Companies such as Chargery in Germany and SparkCharge in the U.S.  have launched mobile charging vehicle services, allowing EV owners to quickly recharge their vehicles in emergency situations \citep{london_mobile_charger_ev_2024}. 
The basic idea is to install a mobile charging station on the truck, and thus, it can travel to the location where the EV is or will be to provide charging services for the EV \citep{cnautonews_ev_chargers_2024}.
NIO in China is a leader in this area, having introduced the 'One Click Power Up' service in 2018, which allows users to order a mobile charging vehicle through their phone from any location. This innovation effectively addresses range anxiety during long-distance trips and offers clear advantages in high-frequency urban travel \citep{xinchuxing_nio_charging_service_2024}.

However, current mobile charging vehicles still require the electric vehicle to stop and wait for charging, which limits their use in extreme emergency scenarios such as medical transportation. For this reason, wireless charging technology has been developed with the aim of utilizing an embedded induction coil system on the road to charge the vehicle in real time while it is in motion. The main challenges of this technology are the high cost of infrastructure and the need for extensive modifications to the existing road system. As a result, the application of the technology is currently limited to pilot projects, such as the wireless charging highway in California\citep{electreon_ers_us_2024}.

\begin{figure}[htbp]
\centering
\begin{minipage}{0.54\textwidth} 
    \centering
    \includegraphics[width=\textwidth]{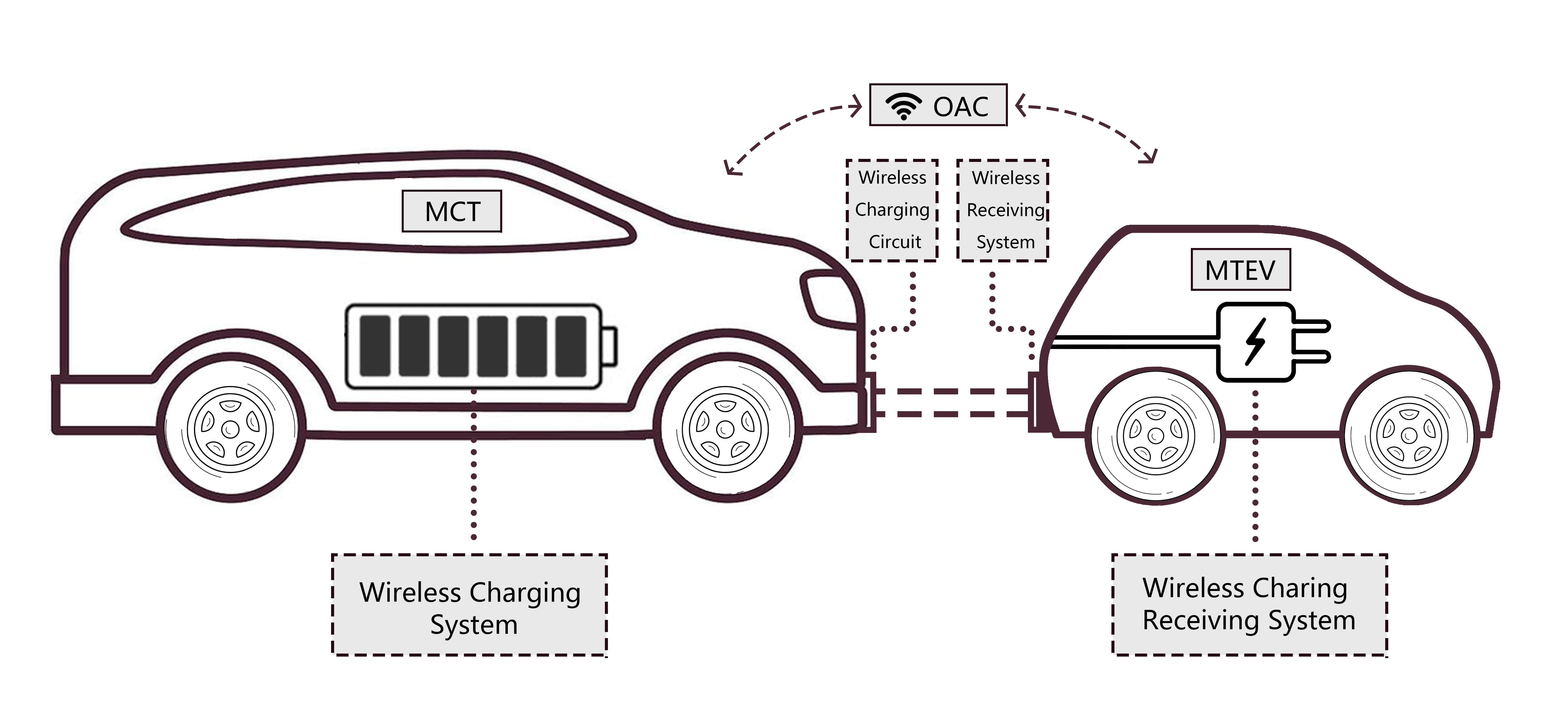}
    \caption{MCT Pattern}
    \label{MCTPattern}
\end{minipage}
\hfill
\begin{minipage}{0.44\textwidth} 
    \centering
    \includegraphics[width=\textwidth]{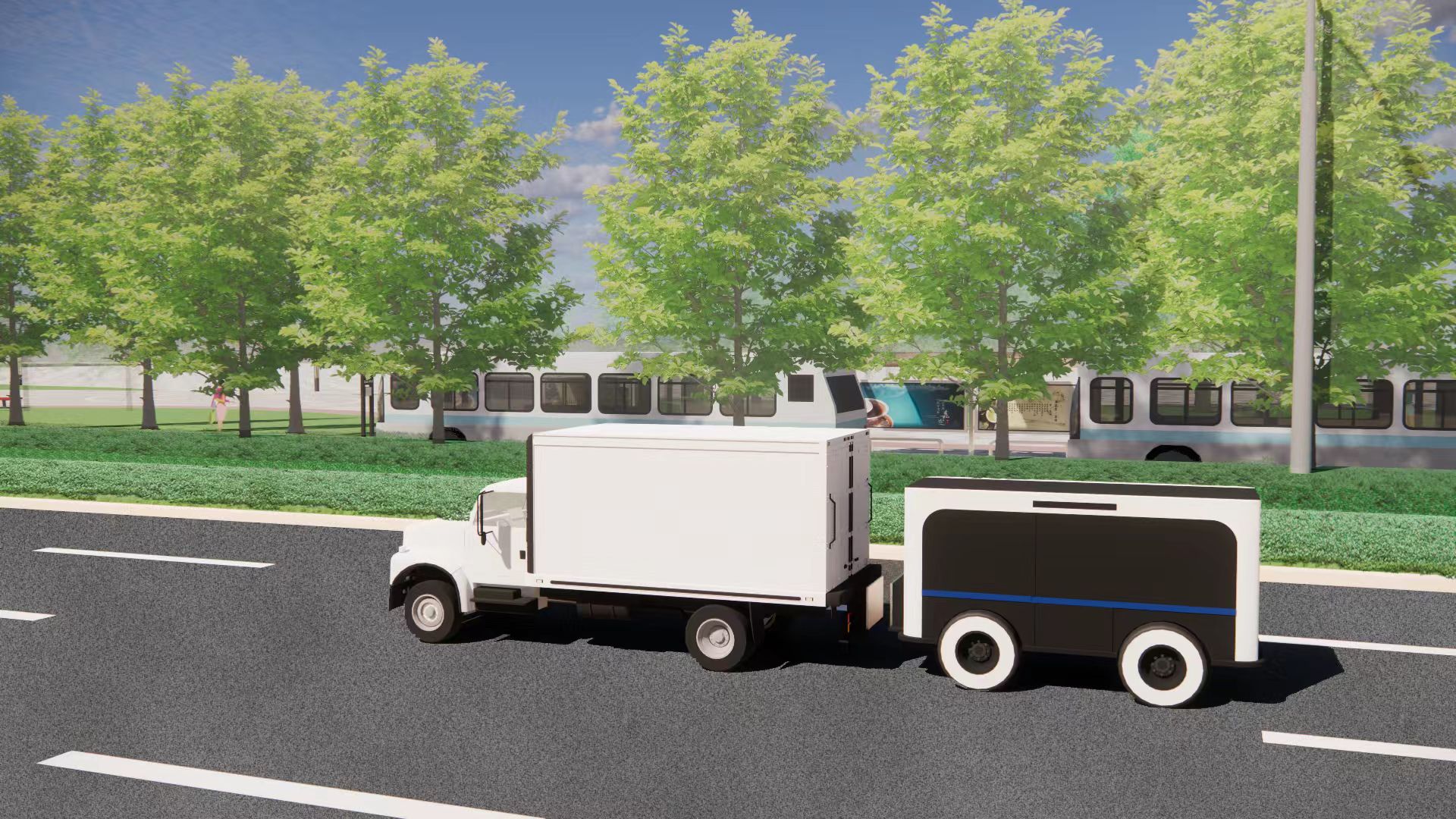}
    \caption{MTEV with MCT on road}
    \label{MCTMTEV}
\end{minipage}
\end{figure}

Unlike the aforementioned technologies, China has developed an innovative approach to wireless charging in-house, inspired by common truck drone systems and in-flight refueling in aviation. Researchers from Northwestern Polytechnical University and Changsha University of Science and Technology, among others, have revealed a mobile wireless charging system for EVs \citep{CN208164783U,CN111002847A,CN220577090U}, shown in Figures~\ref{MCTPattern} and~\ref{MCTMTEV}. These studies focus on the field of EV charging, where vehicles that require charging communicate with power-supplying vehicles through a control center. The charging and receiving sides are equipped with electronic systems that include transmission and reception coils, while the control center manages energy transfer through inductive coupling between the coils, as shown in Figure~\ref{MCTPattern}. This system enables EVs to charge while in motion, avoiding the need to pull over or stop during emergencies.
Recently, Chinese EV companies have been experimenting with this wireless charging method (Figure~\ref{MCTMTEV}), that is, installing the receiving coil on the EV that needs to be charged, and installing the charging system and transmission coil on the truck.
As long as the two coils are detected within a certain distance of each other, they automatically transmit energy wirelessly, so the EV can be charged while driving.
This marks a historic breakthrough in achieving charging while driving.

This paper is inspired by the Organ Transportation Problem (OTP), where an Organ Administration Center (OAC) needs to transport organs to multiple hospitals. Since organ transportation has strict temperature conditions and time windows, the transportation time needs to be minimized. 
We replace the current fixed charging station charging method with the more promising mobile wireless charging method to charge the EV without interrupting the transportation schedule. 
To provide better management insights for the future market, in this paper, we will discuss the Wireless Mobile Charging Electric Vehicle Routing Problem (WMC-EVRP) with the goal of minimizing total costs.
Specifically, at the beginning of each day, the OAC deploys multiple medical transport electric vehicles (MTEVs)  and multiple mobile charging trucks (MCTs), each traveling independently. 
To ensure smooth operation, MCT equipped with wireless charging systems must arrive first and wait for the MTEV at the departure node on the edge to be recharged without delay.
It is worth noting that this would be the coupling that makes the problem exceptionally complex in the study of creatively choosing edges for charging rather than points in the EVRP problem, where the decision-making needs to take into account the need to ensure that the MTEVs are sufficiently charged and the total distance traveled is as short as possible while using as little of the mobile charging distance of the MCTs as possible.

Therefore, to solve this problem, we first propose its mathematical model and a tailored meta-heuristic approach to solve it. Our contributions are threefold.
\begin{enumerate}
\item We propose The use of wireless mobile charging while driving has been proposed for the first time in the electric vehicle routing problem; this technology has been theorized and piloted, and the purpose of this study is to provide management insights for the future market in advance. It is a challenging problem mainly due to the interdependence problem caused by the temporal and spatial synchronization requirements between the two vehicle types (MTEV and MCT).
\item We propose a novel Bitmask Dynamic Programming (BDP) method to address the complexity of determining the optimal charging strategies in this WMC-EVRP. 
This approach allows us to effectively reduce the computational complexity of evaluating numerous potential charging sequences along delivery routes. 
Given each MTEV's route, the BDP algorithm uses binary states to represent whether each edge traversed by the MTEV is charged, utilizing the computer's efficient handling of binary variables to reduce memory usage and processing time.
In this way, we can effectively reduce the computational complexity of evaluating the set of many potentially charging edges on a delivery route.
\item The BDP algorithm is then embedded in the Large Neighborhood Search (LNS) - Local Search (LS) framework to explore potential routes for MTEV. 
Specifically, custom operators designed for the WMC-EVRP are the Charge Removal (CR) and Charge Insertion (CI) operators, which manage the battery capacity of MTEV during the route optimization process.

\item  We present the  
Mixed Integer Linear Program (MILP) for this complex problem and presented the model on a commercial solver, Gurobi. 
The results show that the LNS-BDP framework and Gurobi obtain the same results for small-scale instances, while for medium-sized data, Gurobi fails to obtain an optimal solution within 24 hours, while our results outperform the upper bound (current best solution obtained by Gurobi) with a much shorter runtime. 
Moreover, we investigate the robustness of our algorithm under different constraints, including variations in battery capacity and relative cost of MCT, thus providing management insights.
To validate the practical applicability of our approach, we applied the LNS-BDP algorithm to a real dataset of hospital locations in Singapore, considering the critical factor of transportation time.
 \end{enumerate}

%% file: Sections/2RelatedWork.tex
\section{Related Work} \label{Section:RelatedWork}

\noindent
\textbf{The Electric Vehicle Routing Problems (EVRP)} have generated a significant portion of VRP in recent years. The unique nature of EVRP is its limited battery constraint \citep{EVRPreview}, which often involves refilling at charging stations or swapping batteries over the planning horizon.

Recent work spans different variants of EVRP, the first extension was proposed by \citep{EVRPTW} that are with time windows. \citep{EVRP} proposed the initial design of pick-up and delivery tasks that consider the vehicle load effect on battery consumption. Another main extension is that considers nonlinear charging functions \citep{EVRPNonLinear}, as EV charging has been proven to be non-linear in reality. Both exact methods and heuristic algorithms have been well-performing in EVRP. \citep{BranchCutHeuristic} presents both approaches to solve the Green VRP by combining a branch-and-cut algorithm to strengthen lower bounds and introduced a heuristic approach based on simulated annealing to determine upper bounds. \citep{BranchCutALNSLS} used branch-and-price to give a benchmark set with bounds computed less than 15 customers and developed hybrid Adaptive Large Neighborhood Search (ALNS) heuristic for larger instances. \citep{EVRPTWExactMethod} developed exact branch-price-and-cut algorithms by exploring four variants of the EVRPTW, focusing on different recharging strategies. heuristics, including nearest neighbor heuristic \citep{NNH}, sweeping algorithm\citep{EVRPTW} \citep{LRPBatterySwap}, local search\citep{EVRPLS} and LNS \citep{EVRPLNS}. In addition, \citep{EVRPpartial} used heuristics embedded in a Simulated Annealing framework considering partial recharges and several recharge technologies. Over the past decade, research has increasingly focused on different recharging methods for electric vehicles. \citep{EVRPswapbatterycarbon} proposed a mixed integer programming model and solved with a four-phased heuristic and a two-phased tabu-search algorithm to solve the problem with battery swapping stations, simplifying the charging process. \citep{TDEVRP1} solves the Time-dependent EVRP (TDVRP) with an iterated variable neighborhood search. Unlike intra-route recharging, \citep{LRPBatterySwap} was the first to propose an MIP model for the battery swapping station in location routing problems, while \citep{BatterySwapEVRPTW} introduces the EVRPTW and Synchronized Mobile Battery Swapping, which leverages mobile battery swapping for EVs in freight distribution. \citep{MultipleOptionsEVRP} investigates a new EVRPTW by integrating multiple recharging options: both partial recharging and battery swapping with an improved ant colony optimization algorithm hybridized, an insertion heuristic and enhanced local search. 

\noindent
\textbf{Mobile Charging Technology } is essential for electric vehicles and requires support from modern materials science, mechanics, electrical, operational, and electric research. Current EV charging technology includes stationary charging stations and battery-swapping systems \citep{BatterySwapforEV}. 
Unlike patrol refueling, one of the biggest challenges of EVs is the slow charging and low battery capacity \citep{ChargingforEV}. Traditional charging stations, both public and private, are commercially used to offer various power levels from standard Level 2 chargers to high-speed DC fast chargers \citep{fastchargingreview} \citep{EVchargestandard}. A single charging during the trip usually takes tens of minutes\citep{ChargeDuration}. Over-crowding in charging stations during peak days and hours could also lead to excessive waiting time and trip delays \citep{chargecrowding}. Battery swapping, though less common, provides a quick alternative by replacing depleted batteries with fully charged ones \citep{BatterySwapinfrastructure}. However, there are more limitations on battery swapping for EVs, including the need for standardized battery packs \citep{standardizedforbatteryswap} and the high capital costs associated with swapping infrastructure.  

\begin{figure}[htbp]
\centering
\begin{minipage}{0.84\textwidth}
\includegraphics[width=1\textwidth]{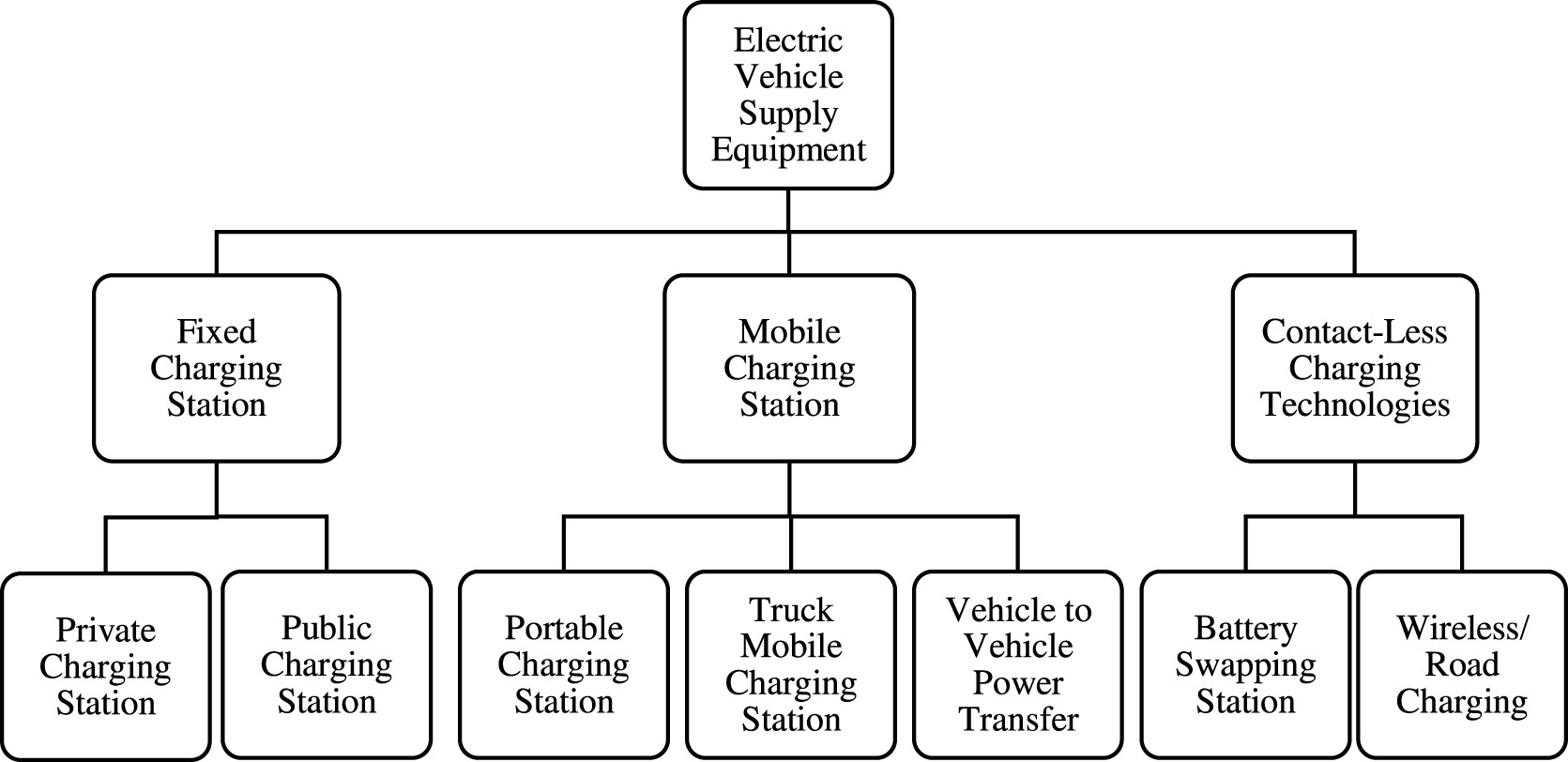}
\label{ChargingMethods}
\end{minipage}
\caption{Different Charging Methods}
\centering
\end{figure}

Recently, advances in mobile charging technology have emerged that offer new possibilities for on-the-go charging solutions \citep{OnTheGoCharge}. The charging lanes \citep{charginglane} or the mobile charging vehicles equipped with high capacity batteries and advanced charging equipment \citep{onroadcharging} can now provide energy to the EVs while both vehicles are in motion \citep{VVChargingNavigation}. This on-the-go mobile charging capability ensures continuous operation without needing EVs to stop charging \citep{wirelesstransfer}. 
In the US, Detroit has unveiled its first wireless charging system on the road, marking a significant milestone in adopting wireless charging on the road \citep{detroitroad}. In China,
more than 40 companies have invested in mobile charging robots \citep{ChinaMobileCharging}. unlike static charging stations, mobile chargers offer greater flexibility by bringing the charger directly to the vehicle, eliminating the need to find an available charging spot.
A collection of EV charging supplies is given in Figure~\ref{ChargingMethods} by \citep{ChargingforEV}.
These innovations significantly improve the flexibility and efficiency of electric vehicle logistics, supporting the latest technological developments in the field. \citep{24hourdelivery} explore the economic viability of battery electric trucks for 24-hour delivery services in city logistics, highlighting the potential cost differences compared to diesel trucks. One of the main reasons why few logistics companies use electric vehicles for deliveries is long operational hours combined with a lack of urgency \citep{EVindelivery}.

\noindent
\textbf{Large Neighborhood Search (LNS) for EVRP}
 have been widely used to solve VRP due to the complexity of the original program. One promising metaheuristic technique that has gained significant attention is LNS, which was first introduced by \citep{LNSfirst} as a local search method, which explores the neighborhoods of the current solution by choosing similar customer visits to destroy from the set of planned routes and repair these routes with a constraint-based tree search.

The versatility of LNS and the variant ALNS has been proven in its application to various VRP and VRP-derived problems with various ruin and reinsert operators introduced \citep{LNS}. Including the frequently used random removal \citep{LNSrandomremoval}, worst removal \citep{ALNSPDTW}, string removal \citep{LNSstringremoval}. \citep{ALNSPDTW} developed an ALNS algorithm for the time-limited pickup and delivery problem, demonstrating the effectiveness of ALNS in solving complex variants of VRP. The emergence of EVs has also led to the study of the EVRP, which introduces additional constraints related to battery capacity and charging requirements. \citep{ALNSFastCharger} proposed an approach that combines LNS with a mixed integer programming model to solve the EVRP with time windows and partial recharging. Similarly, \citep{FlexDelivery} introduced a hybrid heuristic algorithm that integrates LNS and Tabu Search to address EVRP with flexible delivery locations. 
In \citep{LNSVRPwithDeliveryOption}, a VRP with multiple Delivery Options per customer is solved with LNS coupled with a Set Partitioning model. 

Furthermore, recent studies have explored the integration of vehicle load into power estimation functions for the EVRPs. \citep{PartialLinear} proposes a partial linear recharging strategy for EVRP, while \citep{EVRPnonlinearCharging} employed LNS with specialized operators to solve the load-dependent EVRP, demonstrating the adaptability of LNS-based approaches to handle complex energy consumption models. Thus, previous works have proven the versatility and effectiveness of LNS-based algorithms in solving VRP and EVRP variants, including those with time windows, environmental considerations, flexible delivery locations, and load-dependent energy consumption. The integration of local search techniques, such as Tabu Search and Variable Neighborhood Search, are usually add-ons that enhance the performance of LNS algorithms, making them valuable tools for my synergies problem setting.

\noindent
\textbf{Research Gap:}  In this paper, we propose a new problem utilizing state-of-the-art charging technology that considers EVs being charged on the move by trucks loaded with vehicles that can be charged wirelessly, thus overcoming the problem of detours or increased travel time due to charging of charging vehicles in the current variants of EVRP. 
In this context, we consider the sum of the travel cost of the charging vehicle and the EV as an objective, i.e., we consider the collaboration of the two vehicles, which significantly increases the problem's difficulty. 
Although there have been a few papers on EVRP that consider this collaborative relationship, this is the first time that the collaboration between a mobile charging during traveling is considered, i.e., we aim to select the edge where charging is performed instead of the node. This innovative approach improves the efficiency of charging in EVRPs and provides feasibility for using EVs in time-critical problems such as medical transportation.

In addition, DP often plays an important role in solving VRPs and their variants to solve some intractable problems, such as the landmark work \cite{EvolutionDPforVRP},  which  proposed a  DP-based split algorithm for solving the shortest path problem, aiming to devide a giant tour without route delimiters to multiple routes.
Further, \cite{geneticmultidepotperiodicVRP} introduces a DP-based PI operator to solve the multi-depot multi-periodic VRP in a hybrid genetic search method.
Subsequently, \cite{HGSDPforMultitripTDVRP} proposed a DP algorithm to solve the problem of how to split a giant tour and decide the departure time of each route to avoid the peak hours in a multi-trip time-dependent VRP.
Inspired by these, we seek an efficient DP method to solve this complex problem.
We have taken advantage of the combination of the computational advantages of computers for binary operations and the binary nature of the problem (i.e., selecting or not selecting an edge for charging), and have once again landmarked the use of advanced algorithms in the field of computer science for the combinatorial optimization problems.
To the best of our knowledge, this is the first time that Bitmask Dynamic Programming has been used in the VRP family. 

%% file: Sections/3ProblemStatement.tex
\section{Formal Problem Definition}\label{Section:ProblemStatement}

The WMC-EVRP is defined over a complete directed graph \(\mathcal{G} = (\mathcal{I}, \mathcal{A})\), where node set \( \mathcal{I} = \{0, 1, \ldots, n, n+1\}\) representing a hospital involved in medical services. 
The organ demand of each hospital \(i\) is \(d_i\), 
the node \(0\) and  the node \(n+1\) represent the departure and destination OAC.
We define \(\mathcal{A}\) to denote the edges and the distance (or time needed to travel) between hospitals \(i\) and \(j\) is denoted by \(c_{ij}\).

A fleet of \(\mathcal{K}\) MTEVs, each capable of carrying a maximum of \(Q\) organs, is available at the OAC at the beginning of each day. 
A fleet of MCTs, denoted by \(\mathcal{B}\), is stationed at the OAC with fully charged batteries. 
 Each MCT can recharge one or more MTEV multiple times, up to a total of \(\beta\) units of time.
The recharging process is on-road, allowing the MTEV to be recharged by carrying an MCT on the way to the next node. The MCT travels independently to nodes and consumes energy in \(\phi\) units per travel distance. 
 Each MCT begins the day with a fully charged battery capable of supporting \(\mathrm{P}\) time units of travel. 
 For every unit of time that the MCT charges the MTEV, it enables the MTEV to travel an additional $\gamma$ unit of distance.
We need to ensure that all supply demands across the nodes are met at the end of the day. When the MTEV visits a hospital node, the hospital's demand must be completely fulfilled. The costs associated with this problem include the unit acquisition cost for each MTEV \(K_v\), the unit acquisition cost for each MCT \(K_c\), and the travel cost for each unit of distance traveled by a MTEV \(K_t\). 

A mathematical formulation of the OTP is presented below. 

Decision variables \(K\) and \(B\) count the number of MTEV and MCT used during the day. State variable \(U_{k}\) is the number of organs taken by the MTEV \(k\) at the start of the day. Throughout the day, \(u_{ik}\) and \(v_{ib}\) denotes the remaining battery level of MTEV \(k\) and MCT \(b\) at node \(i\). 
\(\delta_{kb}^{ij}\) is the binary variable represent whether MCT \(b\)  charges the MTEV \(k\) on the edge \( (i,j) \in \mathcal{A}\). 
Meanwhile, \(y_{ik}\) and \(w_{ib}\) show whether  \(k\) and \(b\) arrive at each node, while \(x_{ijk}\) and \(z_{ijb}\) tell whether they transport an edge. Finally, \(tk_{ik}\) and \(tb_{ib}\) each denote the time a MTEV or MCT arrives at a node \(i\). The function to be minimized is:
\[
\min \kappa_t\sum_{(i,j)\in \mathcal{A}}\sum_{k\in K}x_{ijk}\tau_{ij} + \kappa_vk + \kappa_cb \tag{1}\label{eq1}
\]
subject to:
\allowdisplaybreaks
\begin{align*}
& m_k \leq \omega, \quad m_k = \sum_{i \in I'} y_{ik} \delta_i, & \forall k \in K \tag{2}\label{eq2}\\
& t_{jk} = \sum_{i \in I} x_{ijk} (t_{ik} + \tau_{ij}), & \forall j \in I, \forall k \in K \tag{3}\label{eq3}\\
& s_{jb} = \sum_{i \in I} z_{ijb} (s_{ib} + \tau_{ij}) + \sum_{i \in I} d_{ijbk} t_{jk}, & \forall j \in I, \forall b \in B \tag{4}\label{eq4}\\
& (\sum_{j \in I} d_{ijkb}) s_{ib} \leq t_{ik}, & \forall i \in I, \forall k \in K, \forall b \in B \tag{5}\label{eq5}\\
& s_{ib} = 0, \quad t_{ik} = 0 & \forall b \in B, \forall k \in K, i = 0 \tag{6}\label{eq6}\\
& u_{jk} = \min (\mathrm{P}, \sum_{i \in I} x_{ijk} (u_{ik} - \tau_{ij} + \gamma \sum_{b \in B} d_{ijkb} \tau_{ij})) , & \forall j \in I, \forall k \in K \tag{7}\label{eq7}\\
& v_{jb} = \sum_{i \in I} z_{ijb} (v_{ib} - \phi \tau_{ij}) + \sum_{i \in I} \sum_{k \in K} d_{ijkb} (v_{ib} - \gamma \tau_{ij}), & \forall j \in I, \forall b \in B \tag{8} \label{eq8}\\
& u_{ik} \geq 0, \quad v_{ib} \geq 0, \quad u_{ik} \leq \mathrm{P}, & \forall i \in I, \forall k \in K, \forall b \in B \tag{9} \label{eq9}\\
& u_{0k} = \mathrm{P}, \quad v_{0b} = \beta & \forall k \in K, \forall b \in B \tag{10} \label{eq10}\\
& \sum_{j \in I^{n+1}} x_{ijk} = \sum_{j \in I^0} x_{jik} = y_{ik}, & \forall i \in I, \forall k \in K \tag{11}\label{eq11}\\
& y_{0k} = y_{(n+1)k} = 1 & \forall k \in K \tag{12} \label{eq12}\\
& \sum_{k \in K} y_{ik} = 1, & \forall i \in I \tag{13} \label{eq13}\\
& \sum_{b \in B} d_{ijkb} \leq x_{ijk}, & \forall i, j \in I, \forall k \in K \tag{14} \label{eq14}\\
& \sum_{i \in I} x_{i0k} = \sum_{i \in I} x_{(n+1)ik} = 0, & \forall k \in K \tag{15} \label{eq15}\\
& \sum_{k \in K} d_{ijkb} \leq w_{ib}, \quad \sum_{k \in K} d_{ijkb} \leq w_{jb} & \forall i, j \in I, \forall b \in B \tag{16} \label{eq16} \\
& \sum_{k \in K} \sum_{j \in I^{n+1}} (z_{ijb} + d_{ijkb}) = \sum_{k \in K} \sum_{j \in I^0} (z_{jib} + d_{jikb}) = w_{ib}, & \forall i \in I, \forall b \in B \tag{17}\label{eq17}\\
& w_{0b} = w_{(n+1)b} = 1 & \forall b \in B \tag{18} \label{eq18}
\end{align*}

Constraint~\eqref{eq2} ensures that the quantity of medical supplies taken by each MTEV \(k\) does not exceed the maximum capacity \(Q\) and that the total demand fulfilled by MTEV \(k\) matches the quantity of supplies it carries.

The Constraint~\eqref{eq3} through Constraint~\eqref{eq6} monitor the travel and arrival time of MTEV and MCT. 
Constraint~\eqref{eq3} calculates the arrival time of an MTEV at node \(j\) based on its departure time from node \(i\) and the travel time between nodes. 
The Constraint~\eqref{eq4} similarly computes the arrival time of an MCT at node \(j\), accounting for both independent travel and travel while attached to an MTEV. Constraint~\eqref{eq5} ensures that if an MCT needs to be connected to an MTEV, it arrives earlier to wait for the MTEV. The initialization Constraints~\eqref{eq6} set the starting times for both MTEV and MCT to zero at the depot.

Constraints~\eqref{eq7} through~\eqref{eq10} handle the energy levels of the MTEV and MCT. Constraint~\eqref{eq7} updates the battery level of an MTEV at node \(j\) from the energy consumed during travel and the energy replenished by MCT. Similarly, Constraint~\eqref{eq8} updates the battery level of an MCT at node \(j\). Constraints~\eqref{eq9} ensure that the battery levels of both MTEV and MCT are non-negative and do not exceed their respective capacities \(\mathrm{P}\) and \(\beta\). The initial conditions for battery levels are set by constraint~\eqref{eq10}.

The Constraints~\eqref{eq11} through~\eqref{eq15} regulate the routes taken by the MTEV. Constraint~\ref {eq11} enforces that the routes form a cycle by equating the number of times an MTEV enters a node to the number of times it exits the node. The Constraint~\eqref{eq12} ensures that each MTEV starts and ends its route at the depot. Constraint~\eqref{eq13} guarantees that each hospital node is visited exactly once by one MTEV. Constraint~\eqref{eq14} ensures that each MCT can only be attached to an MTEV if that MTEV is traversing the corresponding edge. Constraint~\ref{eq15} prohibits MTEV from starting or ending their routes at nodes other than the depot.

The Constraints~\eqref{eq16} through~\eqref{eq18} manage the paths of the MCT. Constraint~\eqref{eq16} ensures that an MCT can only assist an MTEV if it visits the corresponding nodes. The flow conservation Constraint~\eqref{eq17} maintains that MCT, similar to MTEV, forms a cycle. Finally, Constraint~\eqref{eq18} ensures that MCTs start and end their routes at the depot, just like the MTEV.

%% file: Sections/4DP.tex
\section{Bitmask Dynamic Programming}
\label{Section:BDP}

In this section, we present a BDP algorithm to address the challenge of determining optimal charging strategies within WMC-EVRP, particularly when it comes to managing battery charge levels during long-trip delivery routes. 
This BDP algorithm is applied after the delivery routes have been established using the LNS method. 
Given each fixed delivery route, the BDP algorithm determines all possible combinations of charging edges, ensuring that the MTEV can complete its route within the battery limits. The algorithm then chooses the optimal charging method for all MTEV.

To this end, we obtain all feasible charging edge choices for each delivery route by the proposed BDP, and the charging solutions that are non-dominated in terms of the set of edges chosen to be charged are discarded.
A special note on dominant charging: for example, the solution of charging along edges 1 and 5 could be considered dominating the charging solution of edges 1, 3, and 5, as the latter contains redundant charging on edge 3.
Considering all these choices together, we further apply the Deep First Search (DFS) with pruning strategies to find the optimal charging decision.
It is important to note that this optimality is for the charging requirements of all MTEV routes combined, i.e., one charging vehicle MCT may charge multiple MTEV while one MTEV may be charged by multiple MCTs. The framework of BDP is given in Algorithm~\ref{BatteryChargeStateBDP}.

\begin{algorithm}[htbp]
\caption{Battery Charge State BDP Algorithm}
\label{BatteryChargeStateBDP}
\begin{algorithmic}[1]
\STATE \textbf{Input:} $r$ (required remaining battery to complete the route), $\tau$ (distance between nodes), $\mathrm{P}$ (maximum battery capacity), $\gamma$ (charge consumption parameter), $l$ (the entire route)
\STATE $L \leftarrow \text{length}(l)$ 
\STATE $f \leftarrow \text{zeros}(2^{L+1}, \text{dtype=int})$ \hfill {\footnotesize \textcolor{gray}{// Initialize BDP battery matrix} }
\STATE $v \leftarrow \text{zeros}(2^{L+1}, \text{dtype=int})$ \hfill \textcolor{gray}{\footnotesize// Initialize visitation states}
\STATE $f(0) \leftarrow \mathrm{P}$ \hfill \textcolor{gray}{\footnotesize// Initialize battery}
\FOR{$e \in [1$, $L-1$]} 
    \FOR{$j \in [0$, $(1 << (e - 1)) - 1$]} 
        \STATE $k \leftarrow j | (1 << (e - 1))$ \hfill \textcolor{gray}{\footnotesize// \(k\) is the twin state of \(j\) with same preceding branch}
        \IF{$v(j) = 1$}
            \STATE \textbf{continue}
        \ENDIF
        \IF{$f(j) < 0$}
            \STATE $f(j), f(k) \leftarrow -\infty$
            \STATE $v(j), v(k) \leftarrow -1$ \hfill \textcolor{gray}{\footnotesize// Mark as infeasible state}
        \ENDIF
        \STATE $f(k) \leftarrow \min(f(j) - \tau(e) + \gamma\tau(e), \mathrm{P})$ \hfill \textcolor{gray}{\footnotesize// Remaining battery if charge on \(e\)}
        \STATE $f(j) \leftarrow f(j) - \tau(e)$ \hfill \textcolor{gray}{\footnotesize// Remaining battery if no charge on \(e\)}
        \IF{$f(j) \geq r(e)$}
            \STATE $v(j) \leftarrow 1$ \hfill \textcolor{gray}{\footnotesize// Mark as feasible}
            \STATE \textbf{append} $j$ \textbf{to} charge\_node
        \ENDIF
        \IF{$f(k) \geq r(e)$}
            \STATE $v(k) \leftarrow 1$
            \STATE \textbf{append} $k$ \textbf{to} charge\_node
        \ENDIF
    \ENDFOR
\ENDFOR
\STATE \textbf{Output:} charge\_node
\end{algorithmic}
\end{algorithm}

\subsection{BDP Process}

For each delivery route, the BDP approach considers charging or not at each node. By evaluating all feasible charging combinations, the algorithm can identify which charging edges are necessary and which are not. The BDP algorithm iteratively builds a solution that is both feasible and efficient by iteratively updating the battery state at each step, capturing the complexities of managing an MCT charge during long-route deliveries. 

\subsubsection{Notation Definitions}
The notations defined in our BDP algorithm are shown below:
\begin{itemize}

    \item \(\gamma\): Battery charge gained per charging period with an MCT as defined in Section~\ref{Section:ProblemStatement}, which is the ratio of the amount of battery charged to the amount of battery consumed while traveling the same length.
    
    \item \(\tau(e)\): The travel time or distance of the edge \(e\) as defined in Section~\ref{Section:ProblemStatement}, which is also the amount of battery consumed to travel along the edge \(e\).

    \item \(\mathrm{P}\): The battery capacity of each MTEV.

    \item \(r(e)\): The remaining MTEV battery is required to complete the route without further charging after traversing the edge \(e\). 
    We calculate all $r(e) = \sum_{i = e+1}^{m} \tau(i) $ through a preprocessing phase that accumulates from back to front, where \(m\) denotes the length of the route.

    \item \(s\):  the state representation in this BDP. 
    Let $s$ represent the state of charge of each edge, encoded in binary form, indicating whether an edge is charged or not.
    Specifically, a value of \(0\) indicates that no charge occurs on the edge, while \(1\) indicates that charge occurs along this edge. 
    
    \item \(f({e,s})\): the cost-to-go function in this BDP. The remaining level of the battery of the MTEV after traversing the edge \(e\), given the charging status at all previous edges in the current state \(s\).
\end{itemize}

\subsubsection{State Transition Function}
\label{Section:BDP(s)tateTrans}
\begin{figure}[htbp]
\centering
\begin{minipage}{0.75\textwidth}
\includegraphics[width=1\textwidth]{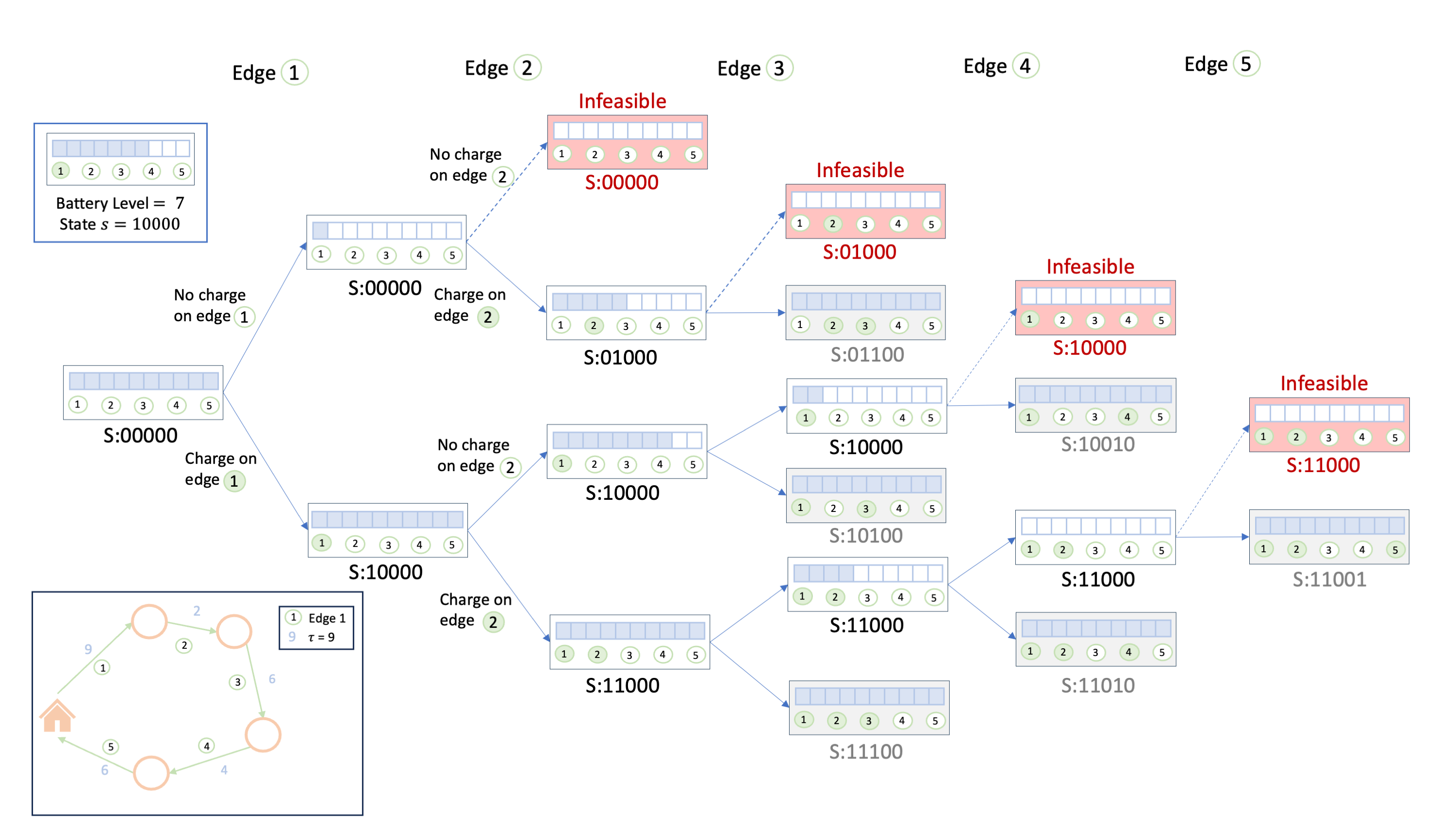}
\end{minipage}
\caption{State Transition and storage of each state}
\label{state_trans1}
\centering
\end{figure}

We define the state transitions used in the BDP algorithm to model the progression of the MTEV along the route. The MTEV can be charged while traversing an edge, with \( e \) representing the current edge and \( e-1 \) representing the preceding edge. Figure~\ref{state_trans1} shows how BDP works along each edge of a route, with the edge numbers and the corresponding energy consumption \(\tau(e)\) labeled in the lower left corner. The blue square represents the battery level of the MTEV, where a full charge is 10 bars. 
Given a known state \(s\) with the remaining battery \(f({e-1,s})\), we update the state for the edge \(e\) considering two possible scenarios: is the MTEV charged by an MCT while traversing the edge or not. 
In Figure~\ref{state_trans1}, the circular green border represents the charging status on each edge: a solid border indicates charging on that edge, while an empty border indicates no charging.

\begin{enumerate}
\item If no charging is taking place on this edge, as shown moving to the next upright branch in Figure~\ref{state_trans1}, the remaining battery after traversing an edge \(e\) is a reduction in energy consumption \(\tau(e)\). The new state \(s'\) for this edge will be established \(s'=s\), and the level of the battery will be reduced by the energy required to traverse the edge \(e\). The transition of the state is shown as follows:
\[
f({e,s'}) = f({e-1,s}) - \tau({e})
\]
\item If the MTEV gets charged while traversing edge \(e\), as shown moving downright branch in Figure~\ref{state_trans1}), the remaining battery is a reduction by energy consumption \(\tau(e)\) and by adding energy charged \(\gamma\tau(e)\) without exceeding battery capacity. In this case, the remaining battery level at state \(s\) is updated to reflect that charging occurs on edge, changing the \(e\)-th bit in the binary representation of \(s\) from \(0\) to \(1\):
\[
f({e,s''})=\min(f({e-1,s})+(\gamma-1)\tau({e}),\mathrm{P})
\]
\end{enumerate}
For edges with only one branch representing the charging case  (e.g., from \texttt{[00000]} to \texttt{[00010]} in 
Figure~\ref{state_trans1}), this means that the MTEV must be charged at the edge to prevent power failure (negative battery level). The gray branches (e.g. \texttt{[00110]}, \texttt{[00101]}, \texttt{[01011]}) in Figure~\ref{state_trans1} represent a state where \(f(e,s)\geq r(e)\), the battery level is sufficient to reach the destination. Therefore, no further charges are needed; we keep the feasible solution.

\begin{figure}[htbp]
\centering
\begin{minipage}{0.75\textwidth}
\includegraphics[width=1\textwidth]{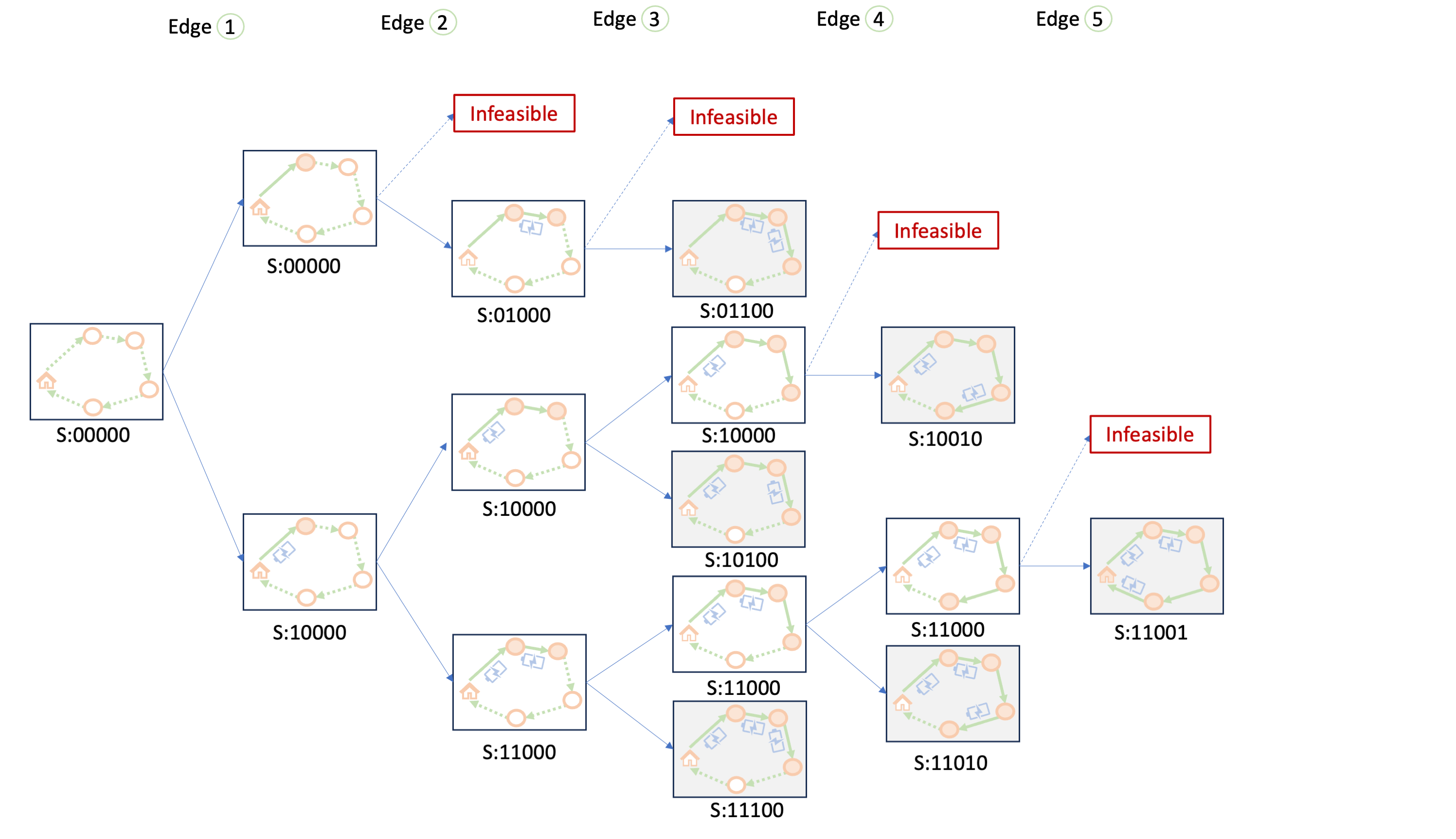}
\end{minipage}
\caption{State Transition on each edge}
\label{state_trans2}
\centering
\end{figure}

We present an example of a 5-edge route scenario shown in Figure~\ref{state_trans2}, starting from \(s=\)\texttt{[00000]}, the initial battery fully charged \(f({0,s}) = \mathrm{P}\). In this figure, dashed lines represent edges that have not yet been visited, while solid lines denote edges already traversed. We begin by evaluating the first edge. If no charging occurs, the updated state is \( s' = \texttt{[00000]} \), and the battery level is given by
\(f({1,s'}) = \mathrm{P} - \tau(0)\). If charging occurs on the first edge, we place a small battery icon on edge one shown in Figure~\ref{state_trans2}. The updated state is \( s'' = \texttt{[00001]} \), with the battery level given by
\(f({1,s''}) = \max(\mathrm{P} + (\gamma - 1)\tau(1), \mathrm{P})\). We then proceed to the next edge. Based on \( s = \texttt{[00000]} \), we have two possible transitions: \( s' = \texttt{[00000]} \) (no charge) and \( s'' = \texttt{[00010]} \) (charge). Similarly, starting from \( s = \texttt{[00001]} \), the transitions are \( s' = \texttt{[00001]} \) and \( s'' = \texttt{[00011]} \).
However, in case of no charge, \(s' = \texttt{[00000]}\), we observe that \( f({2,s'}) = f({1,s}) - \tau(1)\ < 0 \) and a power failure occurs, so this route is marked as infeasible, and all the states derived from \(s'\) (those that retain \( 0 \) in the first and second bits, such as \( s = \texttt{[00100]} \) and \( s = \texttt{[01000]} \)) are also deemed infeasible and excluded from further evaluation. At each stage, infeasible routes are identified and removed if MTEV results in a power failure.
Therefore, as shown in Figure~\ref{state_trans2}, the state \texttt{[00000]} has only one proceeding state \texttt{[00010]}. This process is repeated for all edges until the last state is evaluated. 

\subsubsection{Space Optimization Process}

We observe that the remaining battery on state \(s\), \(f({e,s})\) depends only on the level of the battery immediately preceding \(f({e-1})\). This key characteristic, which reveals the Markov property of the state transitions, allows us to significantly reduce the space complexity of our BDP formulation. Specifically, we can reduce the original two-dimensional state storage to a one-dimensional structure, substantially lowering the memory requirements. In the update process, the non-charging battery \(f({e,s'})\) and the charging battery \(f({e,s''})\)) both depend on the battery of previous state \(s\), \(f({e-1,s})\). By combining the non-charging states before and after traverse \(e\), with battery level \(f({e-1,s})\) and \(f({e,s'})\) into a single variable \(f({s'})\), and always updating the charging state first, followed by the non-charging state, we can ensure the correctness of the transitions. 

To be more specific, we have two possible transitions, as discussed in Section~\ref{Section:BDP(s)tateTrans}. For the scenario where charging does not occur on edge \(e\), the state transition shall be simplified as:
\[
f({s'}) = f({s}) - \tau({e})
\]
Where \(s'=s|0\), which means keeping the original state after this iteration. Given a state before visiting the third edge, \(s=\)\texttt{[10000]} and corresponding battery level \(f(s)\), we update the no-charge state of the third edge by keeping \(s'=\)\texttt{[10000]} and updating the new \(f(s')=f(s)-\tau(3)\).  On the other hand, we update the simplified state in which charging occurs in \(e\) as:
\[
f({s''})=\min{(f(s)+(\gamma-1)\tau(e), \mathrm{P})},
\]
where \(s''=s|(1<<(e-1))\), representing the \((e-1)\)-th bit of \(s\) transformed to \(1\). 
Given the same state \(s=\) \texttt{[10000]} as above, we update the charging state of the third edge by setting \(s''=\)\texttt{[10100]}, and \(f(s''))=\min {(f(s)+(\gamma-1)\tau(e))}\). Similarly to Section~\ref{Section:BDP(s)tateTrans}, this takes into account battery consumption \(\tau(e)\) from state \(s\) to \(s''\), as well as the battery charged \(\gamma\tau(e)\) on edge \(e\), while restricting the maximum battery level \(\mathrm{P}\). 

\begin{figure}[htbp]
\centering
\begin{minipage}{0.75\textwidth}
\includegraphics[width=1\textwidth]{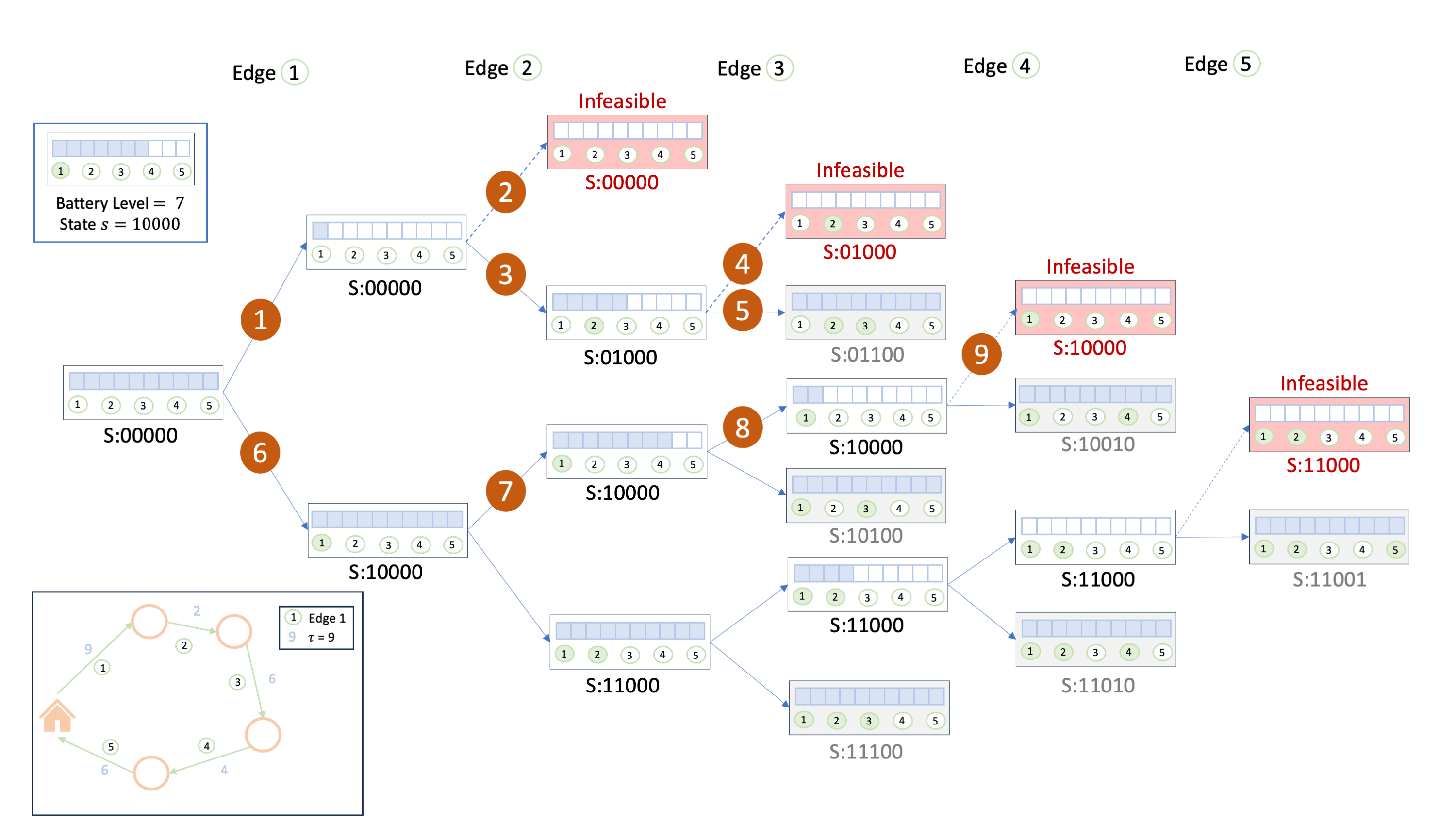}
\end{minipage}
\caption{State Iteration Order on Deep-First Search}
\label{state_trans3}
\centering
\end{figure}

We take the same 5-edge route example in Section~\ref{Section:BDP(s)tateTrans}, and starting from \(s\)=\texttt{[00000]}, we traverse the nodes as far along a route (branch) as possible before it is determined feasible or infeasible. We label the sequence of state visitation in Figure~\ref{state_trans3}, the 1-st visit is to state \texttt{[00000]}, updating \(f(\texttt{[00000]})=\mathrm{P}-\tau(1)\). Similarly, our 2-nd visit updates \(f(\texttt{[00000]})=\mathrm{P}-\tau(1)-\tau(2)\), and we find that \(f(\texttt{[00000]})<0\), so we mark the state \texttt{[00000]} as infeasible as in Figure~\ref{state_trans3}. The 3-rd visit is to charge on the second edge, from \texttt{[00000]} to \texttt{[01000]}, we update the state \(f(\texttt{[01000]})=\min(f(\texttt{[00000]})+(\gamma-1)\tau(2), \mathrm{P})=\min(\mathrm{P}-\tau(1)+(\gamma-1)\tau(2), \mathrm{P})\). Following that, we update the 4-th visit to state \texttt{[01000]}, and determine that \(f(\texttt{[01000]})=\min(\mathrm{P}-\tau(1)+(\gamma-1)\tau(2), \mathrm{P})-\tau(3)<0\), thus label the state \texttt{[01000]} as infeasible. Our 5-th visit is from \texttt{[01000]} to \texttt{[01100]}, we found from the update that \(f(\texttt{[01100]})=\min(f(\texttt{[01000]})+(\gamma-1)\tau(3),\mathrm{P})=\min(\min(\mathrm{P}-\tau(1)+(\gamma-1)\tau(2), \mathrm{P})+(\gamma-1)\tau(3),\mathrm{P})\geq r(3)\), which means the remaining battery at state \texttt{[01100]} is enough for finishing the entire route without further charging, so we label the state \texttt{[01100]} as feasible as in Figure~\ref{state_trans3}. We continue the iteration until each state is marked as either feasible or infeasible, and we give the output of all feasible charging combinations.  

\subsection{Result Storage and Pruning}
\subsubsection{State Marking}
\label{Section:StateMarking}
When solving DP and exploring all possible charging strategies, it is common to find redundant solutions. These redundant solutions do not increase the diversity of the solution set. For example, in the case of charging node sets \(\{1,3,5\}\) and \(\{1,3\}\) with state \texttt{[10101]} and \texttt{[10100]} respectively, the latter can be considered a subset of the former and is therefore regarded redundant. In other words, the charge on the edge \(5\) is unnecessary. To avoid processing these redundant solutions during the iterative process, we introduce a state marking array, denoted as \(v(s)\), to record the validity of each state that works as follows:
\begin{itemize} 
\item \(v(s)=-1\) indicates that the state is infeasible and no further traverse is needed. Examples would be the 2-nd or 4-th visit to \texttt{[00000]} and \texttt{[01000]} with states marked in red in Figure~\ref{state_trans3}. 
\item \(v(s)=0\) indicates that the validity of the state is unknown and that we need to proceed to the next edge. Examples would be the 3-rd or 7-th visit to \texttt{[01000]} and \texttt{[10000]} with states marked in black in Figure~\ref{state_trans3}. 
\item \(v(s)=1\) indicates that the state is feasible and that the MTEV will reach the destination without additional charging. An example would be the 5-th visit to \texttt{[01000]} with state marked in gray in Figure~\ref{state_trans3}. 
\end{itemize}
Initially, all states are marked with a value of \(0\), indicating that their validity is yet to be determined. Our framework can efficiently prune redundant states by implementing this state-marking system, thereby improving the computational efficiency of the BDP process. This ensures that our output contains only unique and valid solutions, reducing unnecessary exploration into deeper branches and focusing on viable charging combinations. When updating the state marking array, we have discussed in Section~\ref{Section:BDP(s)tateTrans}, there are three scenarios to consider for edge \(e\):

\textbf{Scenario 1: Propagation of an Invalid State}

If the state \(f(s)\) has already become negative, it indicates that the state is invalid. Consequently, all subsequent updates from this state will also be invalid. In this case, we mark the state in the array as \(v(s) = -1\), indicating that the state is invalid. This ensures that in further iterations any transitions originating from this invalid state are recognized as invalid, and the algorithm skips over them.

\textbf{Scenario 2: Confirmation of a Valid State}

If the state \(f(s)\) satisfies condition \(f(s) \geq r(e)\), this implies that the MTEV has enough battery to reach the destination without requiring further charging. In this scenario, the state is updated to a valid state by setting \(v(s) = 1\). The valid state is then recorded in the array of feasible solutions.

\textbf{Scenario 3: Pruning of Redundant Solutions}

If during the transition from state \(s \rightarrow s'\) or \(s \rightarrow s''\), the state \(s'\) or \(s''\) has already been marked as valid, then \(s'\) or \(s''\) is considered a redundant solution. In this case, the algorithm directly skips this state, and the feasible state array is not updated with this redundant solution.

\subsubsection{Redundant Solution Removal}

During the update process, the state marking mechanism in Section~\ref{Section:StateMarking} can only omit and remove redundant solutions that have a direct transition relationship such as state \texttt{ [10101]} and redundant state \texttt{[10100]}. However, after obtaining the solution array, there still exist redundant solutions without direct transition relationships, such as charging on edges\({1,3,5}\) and \({3,5}\), that is, state \texttt{ [00101]} and redundant state \texttt{[10101]}. To address this issue, we adopted an additional pruning strategy to remove these redundant solutions, further enhancing the overall performance of the algorithm.

Specifically, we perform a complete traversal of all identified feasible solutions, with a time complexity of \(O(n^2)\), ensuring that each solution is considered. We then use a union operation \(a\cup b\) to determine whether any two states \(a\) and \(b\) have an inclusion relationship. If \(a\cup b = a\), this indicates that \(b\) is a superset of \(a\). In this situation, we choose to prune the larger state \(b\) and retain the smaller subset state \(a\), because the latter represents the most efficient charging strategy with fewer nodes.

This pruning strategy not only reduces the size of the state space but also eliminates unnecessary calculations within the broader LNS framework, hence improving the time efficiency of the algorithm. The dynamic programming algorithm can more effectively and efficiently identify viable solutions by ensuring that only the most essential solutions are retained.

\subsection{Time Complexity}

The time complexity for a single route is \(O(2^{L-1})\), where \(L\) represents the length of the route. This implies that the algorithm needs to evaluate \(2^{L-1}\) possible states, as each node can have two options: to charge or not to charge. The space complexity for this scenario is \(T(2^{L+1})\), as the algorithm needs to store the battery information for each state.

When considering all possible routes, the time complexity becomes \(O\left(\sum_{i=1}^{k} 2^{L_i-1}\right)\), where \(k\) is the number of routes, and \(L_i)\) represents the length of the \(i\)-th route. This indicates that the algorithm needs to evaluate all possible states for all routes. The total space complexity is \(T\left(\sum_{i=1}^{k} 2^{L_i+1}\right)\), reflecting the space required to store the states for all routes.
An alternative to DP is the naive traversal approach to determining charging decisions. This involves iterating through each possible arc and evaluating whether or not to charge at each node. This method evaluates all potential combinations of charging and non-charging states, resulting in a combinatorial explosion of possibilities. The time complexity for this approach is typically \(O(2^n \cdot n!)\), where \(n\) is the number of nodes. This complexity arises because the method must explore all permutations of node sequences, making it infeasible for larger instances.

\begin{figure}[htbp]
\centering
\begin{minipage}{0.64\textwidth}
\includegraphics[width=1\textwidth]{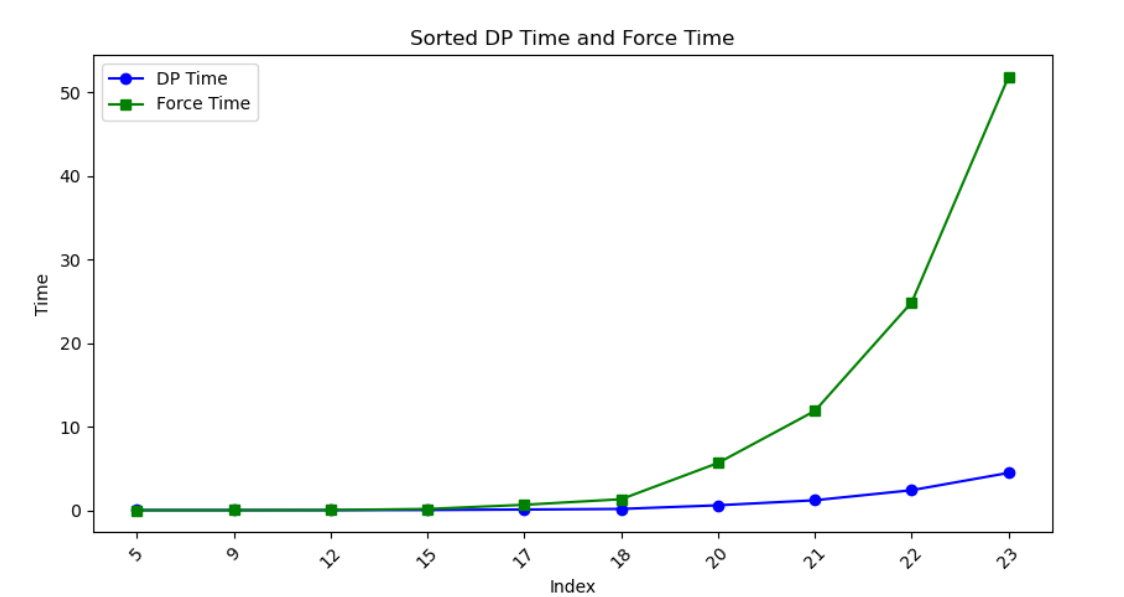}
\label{DPtime}
\end{minipage}
\caption{DP and Naive Traversal Running time}
\centering
\end{figure}

In contrast, our Bitmasking DP approach offers a more efficient solution as we leverage the power of bitmask to represent states. Each state in the DP represents a particular configuration of charging decisions and remaining battery levels, allowing for efficiently reusing previously computed results. Our approach reduces redundant calculations by storing the results of sub-problems, thus avoiding the repeated evaluation of the same state. The time complexity for our approach is \(O(n \cdot 2^n)\). A comparison between the two algorithm complexities has been presented in Figure~\ref{DPtime}. 

%% file: Sections/5Algorithm.tex
\section{LNS for Delivery Vehicle Route} \label{Section:Algorithm}
\begin{figure}[htbp]
\centering
\begin{minipage}{0.55\textwidth}
\includegraphics[width=1\textwidth]{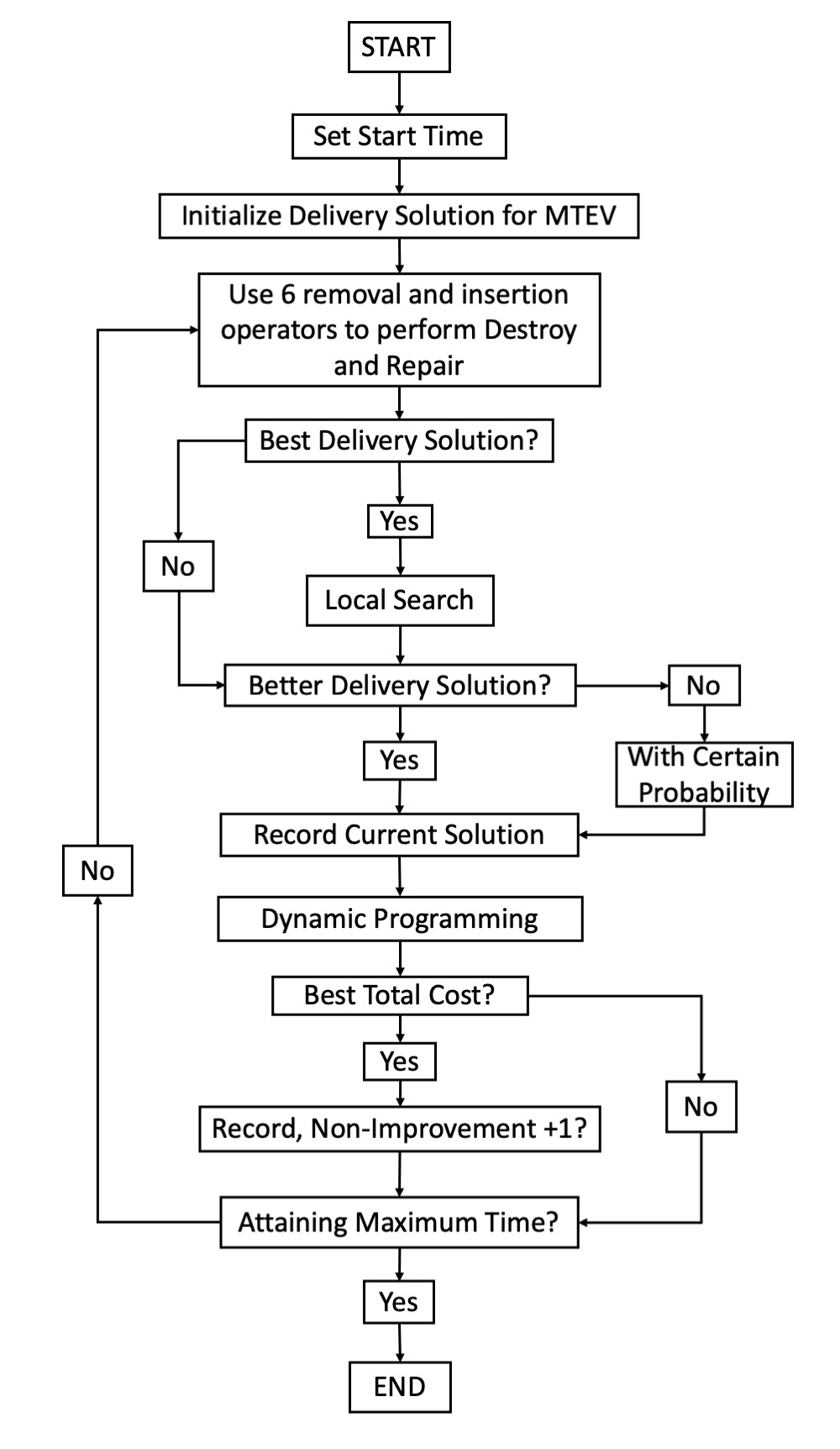}
\label{Algoflow}
\end{minipage}
\caption{The algorithm flow for our problem.}
\centering
\end{figure}

The LNS-LS algorithm focuses on optimizing the routes of MTEV. This method uses a two-tier search strategy to efficiently explore and exploit the solution space. The first tier involves six insertion and six removal operators, rearranging by destroying and repairing the routes, ensuring immediate and precise enhancements. The second tier consists of LS operators that exploit well-performing routes and explore locally to find potential better solutions. A pivotal innovation in our LNS methodology is the Charge Removal (CR) and Charge Insertion (CI) operators, specifically designed for mobile charging routing problems. The CR-CI pair is integral to managing the battery capacities of delivery EVs during route optimization. The complete flow of our LNS-LS framework is given in Figure~\ref{Algoflow}. 


\subsection{The LNS Operators:}

In the destroy and repair step of the algorithm as shown in Figure~\ref{Algoflow}, we use six distinct removal operators to break the current solution, facilitating the exploration of new and improved solutions with six insertion operators to approach a better solution space quickly.  


\begin{enumerate}
\item \textbf{Random Removal (RR):} A set of nodes is randomly selected and removed from the current solution. 
    
\item \textbf{Distance-based Removal (DR):}The nodes that increase the distance of the route most significantly are selected for removal. 

\item \textbf{String Removal (SR):} A segment of the route is removed, the length of the segment to be removed and the starting position of the segment are chosen randomly. 

\item \textbf{Worst Removal (WR):} The nodes that contribute the most to the overall cost are removed. 

\item \textbf{Shaw Removal (ShR):} operator removes the most similar nodes in terms of their location and demand. 



\item \textbf{Random Insertion (RI):} For each removed customer, a random route is selected, and put at the best position within that route with lowest insertion cost. 
\item \textbf{Greedy Insertion (GI):} For each removed customer, identify the position within the current route that results in the lowest increase in total cost. 


\item \textbf{Sequential Insertion (SI):} Insert each removed customer node into the existing routes in a sequential manner.

\item \textbf{Regret-2 Insertion (R2I):} An enhanced method comparing to RI, with a regret function considers not only the best insertion position for each customer but also the second-best option. 
\item \textbf{Regret-3 Insertion (R3I):} Extends the concept of regret insertion by considering the third-best option as well, further enhancing global optimization in terms of potential future impact. 

\item \textbf{Charge Removal (CR)} and \textbf{Charge Insertion (CI)}:

Unlike the other five removal and five insertion operators, CR and CI are always execute in pairs. That is, a CI step is always followed by a CR step. The CR-CI operators are specifically designed for our WMC-EVRP model, which requires managing the battery capacities of delivery EVs during route optimization. We found that, without the presence of CR and CI, it's hard for LNS to allocate a separate MTEV to avoid the use of an MCT. This pair of operators ensures that no vehicle exceeds its battery capacity by removing and reinserting nodes based on their impact on energy consumption.

The CR operator identifies and removes nodes that cause a vehicle's route to exceed its battery capacity, calculating the remaining charge for each route and removing the node that most significantly contributes to energy consumption beyond the vehicle's capacity. 
The CI operator reinserts the removed nodes into the solution while ensuring that each vehicle's path remains within its battery capacity. 
The CR and CI operators are always processed in pairs to ensure that each vehicle operates within its battery constraints while optimizing the overall cost. The CI operator allows us to open new routes to avoid the high cost of wasting unnecessary MCT.
\end{enumerate}

\subsection{Local Search (LS)}

The LS procedure, as shown in Figure~\ref{Algoflow} is the step after destroy and repair, which is used only when we have the current best solution from LNS. We used six LS operators to further explore a small but promising area in the search space, expecting to improve the current solution and find a local optimal. The operators of the LS framework are described as follows:

\begin{enumerate}

\item \textbf{\textit{2-opt Exchange:}} For a pair of edges, remove both of them and reconnect by reversing the segment between them. 

\item \textbf{Or-opt:} Select a pair of consecutive customers from a route and attempt to reinsert them in different positions within the original or different routes.

\item \textbf{2-opt Exchange Multiple Routes:} Select two routes, cut them at some points, and swap the segments between the points of each route. 

\item \textbf{Relocate:}Selects a customer from one route and attempts to insert it into the best position in another different route.

\item \textbf{Exchange:} Select a customer from one route and swaps it with a customer from another route. 

\item \textbf{Cross Exchange:} Selects two consecutive customers from one route and swaps them with two consecutive customers from another route.

\end{enumerate}

After performing LS, we record our current solution of MTEV delivery routes from the LNS-LS framework and feed them into the DP algorithm discussed in Section~\ref{Section:BDP} to identify our optimal charging combination. The iteration stops when the complete LNS-LS-DP framework attains maximum non-improvement counts and outputs the best delivery route for each MTEV and charging route combination of each MCT. 



%% file: Sections/6Experiment.tex
\section{Computational Experimentation}
\label{Section:Experiment}
We run a list of experiments to test the accuracy, robustness, and fitness to the real world. Above all, we would like to know whether our LNS-LS-DP framework is sufficient to combat the WMC-EVRP by providing the correct global optimal value. Another question of our interest is the relative performance of each operator of our LNS-LS, whether or not we could simplify our framework by disabling one or more operators, and how each of them contributes to optimality. Since we are the first to propose the WMC-EVRP and allow synergy between vehicles, we investigate the impact of different MTEV battery capacities and MCT call costs on the overall cost, and we hope to anchor an industrial objective that improves efficiency from the upstream. Finally, we test our algorithm in a Singaporean instance, targeting to find the optimal solutions and confirm our algorithm fits in a real-world setting. The locations of hospitals and depots are randomly generated using a uniform distribution across an Euclidean space, which mimics the geographical randomness of real-world locations. Similarly, the demand at each hospital node is generated independently, with values ranging from 1 to 3 units, representing typical variations in daily or emergency demand. This methodical approach to data generation is designed to ensure that the performance insights we gain are reliable and applicable across various settings. Furthermore, by generating multiple test instances for each complexity level, we aim to provide a comprehensive analysis that helps to understand the scalability and adaptability of the algorithm. 


\subsection{Evaluation of LNS-BDP on Small-Scale Data}

In this subsection, we randomly generate the test instance WMC-EVRP. The raw data specifies the distance matrix between each depot and hospital node and the demand vector for all hospitals. We subsample small instances with 5, 9 and 11 customers and solve them using both our LNS-BDP algorithm, as well as the Gurobi Optimizer, one of the fastest solvers available. We show the results in Table~\ref{tab:Gurobi}, where Gurobi is given an unlimited time for each test instance.
\input{table/Gurobi}
Table~\ref{tab:Gurobi} presents a performance comparison between our LNS-BDP algorithm and the Gurobi solver on small-scale instances. Our LNS-BDP algorithm consistently achieves either the \emph{same optimal solutions} with the Gurobi solver. In some cases, although Gurobi is given a maximum time of 48 hours, it fails to confirm the optimal solution due to the complexity of the problem. In contrast, our algorithm succeeds in finding optimal solutions, requiring significantly less computation time. The table reports both the runtime and the objective values of the best solutions. For each instance, we present the objective cost ($W$) obtained by the Gurobi solver, specifying whether it is the optimal value or the best-found feasible solution (O/F). Our LNS-BDP algorithm runs 10 times for each instance, and we report the best solution found, the average solution value ($W_{\mathrm{avg}}$), and the average runtime ($T(s)$). For each best solution, we include its objective cost ($W_{\mathrm{best}}$) and calculate the gap ($\mbox{Gap} = \left( \frac{W_{\mathrm{best}}}{W} - 1 \right) \times 100\% $)
between the best solution found by the Gurobi solver and the best solution found by our LNS-BDP algorithm.

The results show that our LNS-BDP algorithm can find high-quality solutions that are competitive with and frequently achieve those found by the Gurobi solver. Furthermore, the flexibility of the LNS-BDP framework allows it to perform significantly faster than Gurobi on more complex datasets and larger problem sizes, demonstrating its robustness and potential for WMC-EVRP.

\subsection{Sensitivity analysis of algorithmic components}
\label{sensitivityLNS}

Our LNS-BDP algorithm consists of six removal and six insertion operators designed to solve the WMC-EVRP. We would like to investigate whether we could simplify the algorithm by removing a subset of our LNS framework without a cost or even improve the overall solution quality. We first examine the relative performance of each individual LNS operator and then disable each operator to measure the gap compared to the full LNS model (see Section~\ref{RelativePerformanceLNS}). We also compare the quality of the solution with and without our LS procedure (see Section~\ref{EffectivenessofLS}).

\subsubsection{Relative performances of LNS operators}

\label{RelativePerformanceLNS}
\input{table/Exp2CompareDisableOperator}
We begin by comparing the average running time, average updates per usage, and average total updates of the LNS operators on datasets of varying sizes ranging from 50 to 170 nodes. The average running time represents the computational cost of using each operator, the average updates per usage indicate the efficiency of each operator, and the total updates provided represent the cumulative contribution of an operator across the test instances.

The average running time for each removal operator is given in Table~\ref{tab:avg_removal_run_times}, and the average running time for each insertion operator is given in Table~\ref{tab:avg_iteration_times}. The average running time of each removal and insertion are approximately linearly correlated to the data size, while RR and RI have the lowest time cost, and ShawR, R2I, R3I are significantly more costly than the other operators. We compare the running time against the average updates per usage in Table~\ref{tab:avg_updates_combined}. We found that ShawR, among all removal operators, has contributed more updates, while all removal operators contribute more updates as data size increases. R2I and R3I, on the other hand, contribute more updates for smaller size instances but drop for larger instances. Noticeably, the average updates for GI outperform R2I and R3I on instance sizes over 150, making them inefficient with a high computational cost.

Additionally, we selected a set of 22 instances, each containing between 15 and 170 customers, to evaluate the performance of our algorithm framework under various subsets of the algorithm. Specifically, we explored the impact of disabling each individual operator within the LNS framework. Each configuration was run for a fixed CPU time of 300 seconds, and the best solution obtained in each case was recorded.

Table~\ref{tab:opt_disable} presents the objective costs for the best solutions and the corresponding gaps when each operator was disabled, compared to the full LNS framework. Positive gaps, indicated in blue, suggest that disabling the operator negatively impacted the solution quality, while negative gaps, shown in green, indicate an improvement when the operator was disabled. The sensitivity performance of each operator is summarized in the last row, where we count the instances in which disabling an operator resulted in better, worse, or equivalent solutions. We found that RR and DR, among the removal operators, and RI and SI, among the insertion operators, play significant roles in improving the algorithm’s performance. While the framework works better more frequently without WR and R2I for large instances. These results highlight the importance of each operator in achieving the best possible outcomes, as no single operator can entirely replace the others without compromising solution quality.

\subsubsection{Effectiveness of the LS procedure}
\label{EffectivenessofLS}
\input{table/Exp2LSgap}

Within our LNS algorithm, we implemented the LS procedure to optimize our routes locally. In this experiment, we run our LNS-BDP algorithm with and without the LS procedure to evaluate its effectiveness. We first record the minimum cost for each instance as presented in Section~\ref{RelativePerformanceLNS}. The performance gap of the algorithm without using LS compared to the full LNS-LS is presented in Table~\ref{tab:Exp2LSgap}. We observe that there is a 25\% to 50\% increase in average cost across all instances when LS is not used. Especially for medium and large instances, the gaps are usually between 40\% and 55\%, which indicates the importance of the LS procedure.

\subsection{LNS Performance under Different Scenario Settings}
\input{table/Exp4ChargeCost}
In this section, we evaluate the performance of our LNS algorithm under different scenario settings by tuning key parameters such as the battery capacity of MTEV and the relative cost of MCT to MTEV. When the battery of MTEV is high, we could expect the binding constraint of MTEV to be its capacity, and no MCT will be needed. On the other hand, if the relative cost of MCT is too high, we could expect to allocate more MTEV instead of using MCT. Energy storage has been a continuous challenge in the fields of materials science and engineering. Improvements in battery management systems and increases in battery density can both enhance the battery capacity of our EVs \citep{batterydensity}. Therefore, we find the optimal number of MTEV and MCT under various settings to examine each one's binding constraints.

\begin{figure}[htbp]
\centering
\begin{minipage}{0.53\textwidth}
\includegraphics[width=1\textwidth]{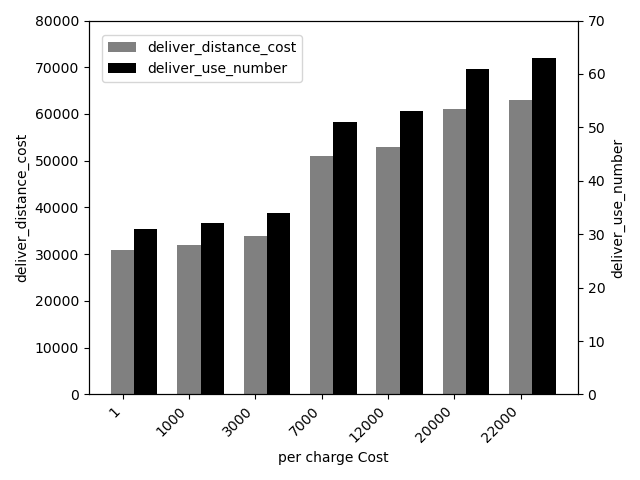}
\end{minipage}
\caption{Comparison of Distance Cost with MCT Cost.}
\label{Exp4ChargeCost}
\centering
\end{figure}

First, we run experiments on various instances with different MTEV battery capacity settings. We present the number of MTEV (\(K\)) and MCT (\(B\)) used, together with the objective costs, in Table~\ref{tab:TruckBattery}. When battery capacity increases, we observe a decrease in MCT demand, but MTEV usage does not increase significantly across each dataset, which is mainly because the optimal routes of the MTEV do not change.
\input{table/Exp3TruckBatteryChange}
Next, we run experiments varying the costs of MCT. The cost of MCT can differ significantly in different parts of the world, and as mobile charging technology advances, the costs of both on-road and off-road chargers may decrease \citep{EVcharging2}. We present our results in Table~\ref{tab:mobilechargercosts}. It is less obvious to identify differences in the number of MTEV and MCT used under most of the MCT cost settings in small datasets. However, we present a graph in Figure~\ref{Exp4ChargeCost} showing the cost for MTEV travel, including the cost of purchasing the MTEV and the distance cost, for a dataset with 300 nodes. We can clearly observe a reduction in the cost per mile as the cost of acquiring MCT decreases. 

\subsection{Experiments on real-world data}

To demonstrate our algorithm for the WMC-EVRP in a real-world setting, we collected the locations of hospitals in Singapore—a developed Southeast Asian nation with advanced medical care and a significant demand for organ transplantation. Singapore has implemented progressive organ donation policies, including the Human Organ Transplant Act \citep{OrganTransAct}, which has evolved over the years to expand eligibility and streamline the organ donation process. This framework supports the high demand for organ transplants in the country \citep{OrganSingapore}, making it well suited for the WMC-EVRP setting.
We selected subsets of hospitals from the collected dataset in Singapore, which included instances with 10, 11, 18, 23, 26, and 29 hospitals. For each instance, we run our algorithm and record the optimal costs, the number of MTEV required, and the number of MCT needed to complete the deliveries efficiently. The results are presented in Table~\ref{tab:real_world_data}.
\input{table/Exp5RealWorld}
The average and best costs for each instance show a clear increase as the number of hospitals grows. This is expected due to the additional complexity and longer routes required to service more hospitals. The fact that the average costs are always close to the best ones suggests that the algorithm consistently finds near-optimal solutions. These real-world examples also gave us an understanding of the hospital number threshold and the travel distance at which MCT becomes necessary. We can also observe that when MCT is needed, the running time increases significantly due to the execution of DP.

%% file: table/Gurobi.tex
\begin{table}[ht]
\caption{Comparison with Gurobi solver on the small scaled data}
\label{tab:Gurobi}
\centering
\resizebox{0.8\textwidth}{!}{
\begin{tabular}{cccccccccc} 
\toprule
\multirow{2}{*}{\textbf{Instance}} & \multicolumn{2}{c}{\textbf{Gurobi}} & \multicolumn{1}{c}{} & \multicolumn{5}{c}{\textbf{LNS-DP}}  \\ 
\cline{2-4} \cline{5-9}
 & \textbf{O/F} & \textbf{W} & \textbf{T(s)} & \textbf{W\_best} & \textbf{Gap} & \textbf{W\_avg} & \textbf{T(s)} \\
\hline
4A & O & 6173 & 2.22      & 6173 & 0     & 6173 & 0.000476 \\
4B & O & 4334 & 0.61      & 4334 & 0     & 4334 & 0.000738 \\
4C & O & 6525 & 3.63      & 6525 & 0     & 6525 & 0.000882 \\
4D & O & 5877 & 2.74      & 5877 & 0     & 5883 & 0.000968 \\
4E & O & 5836 & 2.99      & 5836 & 0     & 5878 & 0.001006 \\
\hdashline
8A & O & 6981 & 5556.12   & 6981 & 0     & 6981 & 0.003147 \\
8B & O & 6957 & 30284.52  & 6957 & 0     & 6957 & 0.002726 \\
8C & O & 6708 & 23110.93  & 6708 & 0     & 6708 & 0.003056 \\
8D & O & 6023 & 734.14    & 6023 & 0     & 6023 & 0.003865 \\
8E & O & 5816 & 89387.50  & 5816 & 0     & 5827 & 0.002822 \\
\hdashline
10A & O & 8461 & 40212.00 & 8461 & 0     & 8465 & 0.005285 \\
10B & O & 7215 & 67636.00 & 7215 & 0     & 7218 & 0.052951 \\
10C & O & 8633 & 63678.00 & 8633 & 0     & 8640 & 0.009749 \\
10D & F & 6923 & 172800.00& 6923 & 0     & 6923 & 0.044072 \\
10E & O & 7769 & 25631.56 & 7769 & 0     & 7802 & 0.004841 \\
\hdashline
12A & F & 7293 & 172800.00& 7293 & 0     & 7300 & 0.790559 \\
12B & F & 8651 & 172800.00& 8651 & 0     & 8779 & 0.009230 \\
12C & F & 9224 & 172800.00& 9224 & 0     & 9257 & 0.043811 \\
12D & F & 8216 & 172800.00& 8216 & 0     & 8218 & 0.049202 \\
12E & F & 8373 & 172800.00& 8373 & 0     & 8373 & 0.030367 \\
\bottomrule
\end{tabular}
}
\end{table}

%% file: table/Exp2CompareDisableOperator.tex
\begin{table}[htbp]
\caption{Comparison of the quality of the best solutions found by each framework.}
\label{tab:opt_disable}
\centering
\resizebox{1\textwidth}{!}{
\begin{tabular}{ccccccccccccc} 
\toprule
\multirow{2}{*}{\begin{tabular}[c]{@{}c@{}}\textbf{Ins}\\\textbf{(}n\textbf{)}\end{tabular}} & \multirow{2}{*}{\begin{tabular}[c]{@{}c@{}}Full\\($W$)\end{tabular}} & \multicolumn{5}{c}{Disabled removal operator} &  & \multicolumn{5}{c}{Disabled insertion operator} \\ 
\cline{3-7}\cline{9-13}
 &  & RR & DR & SR & WR & ShR &  & RI & GI & SI & R2I & R3I \\ 
\hline
\textbf{15A} & 8244 & 8230 & 8260 & 8230 & 8260 & 8230 &  & 8230 & 8230 & 8260 & 8230 & 8260 \\
 &  & \textcolor[rgb]{0,0.502,0}{14} & \textcolor{blue}{-16} & \textcolor[rgb]{0,0.502,0}{14} & \textcolor{blue}{-16} & \textcolor[rgb]{0,0.502,0}{14} &  & \textcolor[rgb]{0,0.502,0}{14} & \textcolor[rgb]{0,0.502,0}{14} & \textcolor[rgb]{0,0.502,0}{14} & \textcolor{blue}{-16} & \textcolor[rgb]{0,0.502,0}{14} \\
\textbf{20A} & 9830 & 9799 & 9783 & 9814 & 9804 & 9799 &  & 9820 & 9814 & 9794 & 9835 & 9743 \\
 &  & \textcolor[rgb]{0,0.502,0}{31} & \textcolor[rgb]{0,0.502,0}{47} & \textcolor[rgb]{0,0.502,0}{16} & \textcolor[rgb]{0,0.502,0}{26} & \textcolor[rgb]{0,0.502,0}{31} &  & \textcolor[rgb]{0,0.502,0}{31} & \textcolor[rgb]{0,0.502,0}{10} & \textcolor[rgb]{0,0.502,0}{16} & \textcolor[rgb]{0,0.502,0}{36} & \textcolor{blue}{-5} \\
\textbf{25A} & 12414 & 12364 & 12047 & 12304 & 12394 & 12221 &  & 12151 & 12226 & 12141 & 13127 & 12417 \\
 &  & \textcolor[rgb]{0,0.502,0}{50} & \textcolor[rgb]{0,0.502,0}{367} & \textcolor[rgb]{0,0.502,0}{110} & \textcolor[rgb]{0,0.502,0}{20} & \textcolor[rgb]{0,0.502,0}{193} &  & \textcolor[rgb]{0,0.502,0}{253} & \textcolor[rgb]{0,0.502,0}{263} & \textcolor[rgb]{0,0.502,0}{188} & \textcolor[rgb]{0,0.502,0}{273} & \textcolor{blue}{-713} \\
\textbf{30A} & 13908 & 14352 & 14594 & 14406 & 14377 & 14272 &  & 14434 & 13931 & 14204 & 15765 & 14216 \\
 &  & \textcolor{blue}{-444} & \textcolor{blue}{-686} & \textcolor{blue}{-498} & \textcolor{blue}{-469} & \textcolor{blue}{-364} &  & \textcolor{blue}{-503} & \textcolor{blue}{-526} & \textcolor{blue}{-23} & \textcolor{blue}{-296} & \textcolor{blue}{-1857} \\
\textbf{35A} & 17054 & 16785 & 17056 & 17066 & 17062 & 17050 &  & 16911 & 17088 & 16559 & 16909 & 16782 \\
 &  & \textcolor[rgb]{0,0.502,0}{269} & \textcolor{blue}{-2} & \textcolor{blue}{-12} & \textcolor{blue}{-8} & \textcolor[rgb]{0,0.502,0}{4} &  & \textcolor[rgb]{0,0.502,0}{7} & \textcolor[rgb]{0,0.502,0}{143} & \textcolor{blue}{-34} & \textcolor[rgb]{0,0.502,0}{495} & \textcolor[rgb]{0,0.502,0}{145} \\
\textbf{45A} & 19430 & 19421 & 19427 & 19429 & 19448 & 19451 &  & 19430 & 19420 & 19446 & 19421 & 19424 \\
 &  & \textcolor[rgb]{0,0.502,0}{9} & \textcolor[rgb]{0,0.502,0}{3} & \textcolor[rgb]{0,0.502,0}{1} & \textcolor{blue}{-18} & \textcolor{blue}{-21} &  & \textcolor{blue}{-10} & 0 & \textcolor[rgb]{0,0.502,0}{10} & \textcolor{blue}{-16} & \textcolor[rgb]{0,0.502,0}{9} \\
\textbf{50A} & 21135 & 21123 & 21107 & 21111 & 21113 & 21111 &  & 21131 & 21137 & 21123 & 21091 & 21111 \\
 &  & \textcolor[rgb]{0,0.502,0}{12} & \textcolor[rgb]{0,0.502,0}{28} & \textcolor[rgb]{0,0.502,0}{24} & \textcolor[rgb]{0,0.502,0}{22} & \textcolor[rgb]{0,0.502,0}{24} &  & \textcolor[rgb]{0,0.502,0}{4} & 0 & \textcolor[rgb]{0,0.502,0}{12} & \textcolor[rgb]{0,0.502,0}{44} & \textcolor[rgb]{0,0.502,0}{24} \\
\textbf{55A} & 24458 & 24873 & 24549 & 24533 & 25279 & 24614 &  & 24314 & 24311 & 24374 & 25157 & 24343 \\
 &  & \textcolor{blue}{-415} & \textcolor{blue}{-91} & \textcolor{blue}{-75} & \textcolor{blue}{-821} & \textcolor{blue}{-156} &  & \textcolor[rgb]{0,0.502,0}{144} & \textcolor[rgb]{0,0.502,0}{147} & \textcolor[rgb]{0,0.502,0}{102} & \textcolor{blue}{-15} & \textcolor[rgb]{0,0.502,0}{115} \\
\textbf{60A} & 24351 & 24231 & 24066 & 24029 & 23942 & 24054 &  & 23812 & 24104 & 24011 & 24409 & 23997 \\
 &  & \textcolor[rgb]{0,0.502,0}{120} & \textcolor[rgb]{0,0.502,0}{285} & \textcolor[rgb]{0,0.502,0}{322} & \textcolor[rgb]{0,0.502,0}{409} & \textcolor[rgb]{0,0.502,0}{297} &  & \textcolor[rgb]{0,0.502,0}{539} & \textcolor[rgb]{0,0.502,0}{207} & \textcolor[rgb]{0,0.502,0}{340} & \textcolor{blue}{-58} & \textcolor[rgb]{0,0.502,0}{354} \\
\textbf{65A} & 26888 & 26577 & 26864 & 26984 & 26755 & 26724 &  & 26885 & 26628 & 26961 & 26674 & 27035 \\
 &  & \textcolor[rgb]{0,0.502,0}{311} & \textcolor[rgb]{0,0.502,0}{24} & \textcolor{blue}{-96} & \textcolor[rgb]{0,0.502,0}{133} & \textcolor[rgb]{0,0.502,0}{164} &  & \textcolor[rgb]{0,0.502,0}{3} & \textcolor[rgb]{0,0.502,0}{260} & \textcolor{blue}{-73} & \textcolor[rgb]{0,0.502,0}{214} & \textcolor{blue}{-147} \\
\textbf{75A} & 27643 & 27588 & 27603 & 27603 & 27608 & 27608 &  & 27431 & 27640 & 27601 & 27578 & 27656 \\
 &  & \textcolor[rgb]{0,0.502,0}{55} & \textcolor[rgb]{0,0.502,0}{40} & \textcolor[rgb]{0,0.502,0}{40} & \textcolor[rgb]{0,0.502,0}{35} & \textcolor[rgb]{0,0.502,0}{35} &  & \textcolor[rgb]{0,0.502,0}{212} & \textcolor[rgb]{0,0.502,0}{3} & \textcolor[rgb]{0,0.502,0}{42} & \textcolor[rgb]{0,0.502,0}{65} & \textcolor{blue}{-13} \\
\textbf{85A} & 32164 & 32128 & 32354 & 32666 & 32451 & 32500 &  & 32655 & 32532 & 32255 & 32212 & 32352 \\
 &  & \textcolor[rgb]{0,0.502,0}{36} & \textcolor{blue}{-190} & \textcolor{blue}{-502} & \textcolor{blue}{-287} & \textcolor{blue}{-336} &  & \textcolor{blue}{-491} & \textcolor{blue}{-368} & \textcolor{blue}{-91} & \textcolor{blue}{-48} & \textcolor{blue}{-188} \\
\textbf{90A} & 35732 & 34682 & 34222 & 34650 & 34115 & 34242 &  & 34389 & 35131 & 35216 & 35193 & 34222 \\
 &  & \textcolor[rgb]{0,0.502,0}{1050} & \textcolor[rgb]{0,0.502,0}{1510} & \textcolor[rgb]{0,0.502,0}{1082} & \textcolor[rgb]{0,0.502,0}{1617} & \textcolor[rgb]{0,0.502,0}{1490} &  & \textcolor[rgb]{0,0.502,0}{343} & \textcolor{blue}{-399} & \textcolor{blue}{-484} & \textcolor{blue}{-461} & \textcolor[rgb]{0,0.502,0}{510} \\
\textbf{100A} & 33448 & 33428 & 33109 & 34590 & 34149 & 34507 &  & 34215 & 34111 & 34124 & 33962 & 34273 \\
 &  & \textcolor{blue}{-20} & \textcolor[rgb]{0,0.502,0}{339} & \textcolor{blue}{-1142} & \textcolor{blue}{-701} & \textcolor{blue}{-1059} &  & \textcolor{blue}{-72} & \textcolor{blue}{-767} & \textcolor{blue}{-663} & \textcolor{blue}{-676} & \textcolor{blue}{-514} \\
\textbf{105A} & 38947 & 39165 & 41168 & 39923 & 39758 & 39632 &  & 41405 & 37854 & 40949 & 40785 & 41755 \\
 &  & \textcolor{blue}{-218} & \textcolor{blue}{-2221} & \textcolor{blue}{-976} & \textcolor{blue}{-811} & \textcolor{blue}{-685} &  & \textcolor[rgb]{0,0.502,0}{1106} & \textcolor{blue}{-2458} & \textcolor[rgb]{0,0.502,0}{1093} & \textcolor{blue}{-2002} & \textcolor{blue}{-1838} \\
\textbf{115A} & 43317 & 39838 & 39946 & 40462 & 38862 & 39578 &  & 43129 & 38485 & 42115 & 41366 & 41942 \\
 &  & \textcolor[rgb]{0,0.502,0}{3479} & \textcolor[rgb]{0,0.502,0}{3371} & \textcolor[rgb]{0,0.502,0}{2855} & \textcolor[rgb]{0,0.502,0}{4455} & \textcolor[rgb]{0,0.502,0}{3739} &  & \textcolor[rgb]{0,0.502,0}{4110} & \textcolor{blue}{188} & \textcolor[rgb]{0,0.502,0}{4832} & \textcolor[rgb]{0,0.502,0}{1202} & \textcolor[rgb]{0,0.502,0}{1951} \\
\textbf{125A} & 44553 & 43062 & 41218 & 41773 & 46127 & 40785 &  & 42529 & 41395 & 41319 & 42776 & 41757 \\
 &  & \textcolor[rgb]{0,0.502,0}{1491} & \textcolor[rgb]{0,0.502,0}{3335} & \textcolor[rgb]{0,0.502,0}{2780} & \textcolor{blue}{-1574} & \textcolor[rgb]{0,0.502,0}{3768} &  & \textcolor[rgb]{0,0.502,0}{353} & \textcolor[rgb]{0,0.502,0}{2024} & \textcolor[rgb]{0,0.502,0}{3158} & \textcolor[rgb]{0,0.502,0}{3234} & \textcolor[rgb]{0,0.502,0}{1777} \\
\textbf{135A} & 50383 & 49835 & 50850 & 51045 & 57560 & 52149 &  & 51527 & 48769 & 52674 & 49123 & 52715 \\
 &  & \textcolor{blue}{-548} & \textcolor[rgb]{0,0.502,0}{-467} & \textcolor[rgb]{0,0.502,0}{-662} & \textcolor{blue}{-7177} & \textcolor{blue}{-1766} &  & \textcolor[rgb]{0,0.502,0}{2108} & \textcolor{blue}{-1144} & \textcolor[rgb]{0,0.502,0}{1614} & \textcolor{blue}{-2291} & \textcolor[rgb]{0,0.502,0}{1260} \\
\textbf{145A} & 52320 & 53444 & 52111 & 52484 & 54907 & 52264 &  & 55876 & 51072 & 53639 & 52294 & 54569 \\
 &  & \textcolor{blue}{-1124} & \textcolor[rgb]{0,0.502,0}{209} & \textcolor{blue}{-164} & \textcolor{blue}{-2587} & \textcolor[rgb]{0,0.502,0}{56} &  & \textcolor[rgb]{0,0.502,0}{2828} &\textcolor{blue}{-3556} & \textcolor[rgb]{0,0.502,0}{1248} & \textcolor{blue}{-1319} & \textcolor[rgb]{0,0.502,0}{26} \\
\textbf{155A} & 57041 & 56718 & 56330 & 56483 & 58985 & 55317 &  & 57194 & 54256 & 57277 & 56872 & 57477 \\
 &  & \textcolor[rgb]{0,0.502,0}{323} & \textcolor[rgb]{0,0.502,0}{711} & \textcolor[rgb]{0,0.502,0}{558} & \textcolor{blue}{-1944} & \textcolor[rgb]{0,0.502,0}{1724} &  & \textcolor[rgb]{0,0.502,0}{1157} & \textcolor{blue}{-153} & \textcolor[rgb]{0,0.502,0}{2785} & \textcolor{blue}{-236} & \textcolor[rgb]{0,0.502,0}{169} \\
\textbf{165A} & 58029 & 59341 & 57631 & 55931 & 62231 & 57251 &  & 57076 & 55406 & 63442 & 57146 & 60656 \\
 &  & \textcolor{blue}{-1312} & \textcolor[rgb]{0,0.502,0}{398} & \textcolor[rgb]{0,0.502,0}{2098} & \textcolor{blue}{-4202} & \textcolor{blue}{778} &  & \textcolor{blue}{-2460} & \textcolor[rgb]{0,0.502,0}{953} & \textcolor[rgb]{0,0.502,0}{2623} & \textcolor{blue}{-5413} & \textcolor[rgb]{0,0.502,0}{883} \\
\textbf{170A} & 59437 & 59092 & 60238 & 65250 & 63786 & 59109 &  & 61667 & 60410 & 62364 & 60651 & 61463 \\
 &  & \textcolor[rgb]{0,0.502,0}{345} & \textcolor{blue}{-801} & \textcolor{blue}{-5813} & \textcolor{blue}{-4349} & \textcolor[rgb]{0,0.502,0}{328} &  & \textcolor{blue}{-1284} & \textcolor{blue}{-2230} & \textcolor{blue}{-973} & \textcolor{blue}{-2927} & \textcolor{blue}{-1214} \\
\textbf{Performance} & \textcolor[rgb]{0,0.502,0}{Worse}/ \textcolor{blue}{Better}/ Same & \textcolor[rgb]{0,0.502,0}{15}/ \textcolor{blue}{7}/ 0 & \textcolor[rgb]{0,0.502,0}{15}/ \textcolor{blue}{7}/ 0 & \textcolor[rgb]{0,0.502,0}{13}/ \textcolor{blue}{9}/ 0 & \textcolor[rgb]{0,0.502,0}{8}/ \textcolor{blue}{14}/ 0 & \textcolor[rgb]{0,0.502,0}{14}/ \textcolor{blue}{8}/ 0 & & \textcolor[rgb]{0,0.502,0}{16}/ \textcolor{blue}{6}/ 0 & \textcolor[rgb]{0,0.502,0}{10}/ \textcolor{blue}{10}/ 2 & \textcolor[rgb]{0,0.502,0}{15}/ \textcolor{blue}{7}/ 0 & \textcolor[rgb]{0,0.502,0}{8}/ \textcolor{blue}{14}/ 0 & \textcolor[rgb]{0,0.502,0}{13}/ \textcolor{blue}{9}/ 0 \\
\bottomrule
\end{tabular}
}
\end{table}

%% file: table/Exp2LSgap.tex
\begin{table}[H]
\centering
\caption{Performance of Algorithm with and without LS Procedure}
\label{tab:Exp2LSgap}
\begin{tabular}{lccccc|lccccc}
\toprule
\textbf{Data} & \textbf{w LS} & \textbf{wo LS} & \textbf{Diff} & \textbf{Gap} & &
\textbf{Data} & \textbf{w LS} & \textbf{wo LS} & \textbf{Diff} & \textbf{Gap} \\
\midrule
60A  & 21802  & 27960  & -6158  & \textcolor{green!50!black}{28.25\%} & &
115A & 37276  & 54337  & -17061 & \textcolor{red!50!black}{45.77\%} \\
65A  & 23458  & 30044  & -6586  & \textcolor{green!50!black}{28.08\%} & &
120A & 39294  & 59002  & -19708 & \textcolor{red!50!black}{46.75\%} \\
70A  & 23650  & 30122  & -6472  & \textcolor{green!50!black}{27.37\%} & &
125A & 39294  & 59112  & -19818 & \textcolor{red!75!black}{50.44\%} \\
75A  & 25586  & 34875  & -9289  & \textcolor{yellow!50!black}{36.31\%} & &
130A & 40584  & 63252  & -22668 & \textcolor{red!75!black}{55.93\%} \\
80A  & 26620  & 36735  & -10109 & \textcolor{yellow!50!black}{37.96\%} & &
135A & 44537  & 64784  & -20247 & \textcolor{red!50!black}{45.46\%} \\
85A  & 27516  & 39687  & -12171 & \textcolor{yellow!75!black}{44.23\%} & &
140A & 42843  & 62531  & -19688 & \textcolor{red!50!black}{45.96\%} \\
90A  & 29520  & 43956  & -14436 & \textcolor{red!50!black}{48.89\%} & &
145A & 45749  & 66386  & -20644 & \textcolor{red!50!black}{45.13\%} \\
95A  & 33246  & 48814  & -15568 & \textcolor{red!50!black}{46.83\%} & &
150A & 53292  & 73407  & -20115 & \textcolor{yellow!50!black}{37.74\%} \\
100A & 32874  & 46627  & -13753 & \textcolor{yellow!75!black}{41.84\%} & &
155A & 54112  & 77412  & -23300 & \textcolor{yellow!75!black}{43.06\%} \\
105A & 36981  & 52557  & -15576 & \textcolor{yellow!75!black}{42.69\%} & &
160A & 51014  & 76127  & -25113 & \textcolor{red!50!black}{49.23\%} \\
110A & 37760  & 53190  & -15430 & \textcolor{yellow!75!black}{41.48\%} & &
165A & 51014  & 76127  & -25113 & \textcolor{red!50!black}{49.23\%} \\
 & & & & & &
170A & 56003  & 74861  & -18858 & \textcolor{yellow!50!black}{33.67\%} \\
\bottomrule
\end{tabular}
\end{table}

%% file: table/Exp4ChargeCost.tex
\begin{table}[H]
\centering
\caption{Objective Costs Under Various Mobile Charger Costs}
\label{tab:mobilechargercosts}
\begin{tabular}{lccccc|lccccc|lccccc}
\toprule
\multicolumn{6}{c|}{\textbf{35A}} & \multicolumn{6}{c|}{\textbf{75A}} & \multicolumn{6}{c}{\textbf{165A}} \\
\(\kappa_B\) & \textbf{Avg} & \textbf{Best} & \(B\) & \(K\) & &
\(\kappa_B\) & \textbf{Avg} & \textbf{Best} & \(B\) & \(K\) & &
\(\kappa_B\) & \textbf{Avg} & \textbf{Best} & \(B\) & \(K\) \\
\midrule
1     & 11084 & 11084 & 2  & 4  & & 1     & 18983 & 18983 & 2  & 8  & & 1     & 36363 & 36363 & 3  & 17 \\
100   & 11322 & 11322 & 2  & 4  & & 100   & 18987 & 18987 & 2  & 8  & & 100   & 36677 & 36677 & 2  & 17 \\
1000  & 13078 & 13078 & 2  & 4  & & 1000  & 20945 & 20945 & 2  & 8  & & 1000  & 38344 & 38344 & 2  & 17 \\
3000  & 17040 & 17040 & 2  & 4  & & 3000  & 24903 & 24903 & 2  & 8  & & 3000  & 42514 & 42514 & 2  & 17 \\
5000  & 21057 & 21057 & 2  & 4  & & 5000  & 28812 & 28812 & 2  & 8  & & 5000  & 46672 & 46672 & 2  & 17 \\
50000 & 32127 & 32127 & 0  & 15 & & 50000 & 32489 & 32489 & 0  & 15 & & 50000 & 59138 & 59138 & 0  & 27 \\
\bottomrule
\end{tabular}
\end{table}

%% file: table/Exp3TruckBatteryChange.tex
\begin{table}[H]
\centering
\caption{Mobile Charger Usage when Truck Battery Changes}
\label{tab:TruckBattery}
\begin{tabular}{lccccc|lccccc|lccccc}
\toprule
\multicolumn{6}{c|}{\textbf{100A}} & \multicolumn{6}{c|}{\textbf{120A}} & \multicolumn{6}{c}{\textbf{150A}} \\
\(\mathrm{P}\) & \textbf{Avg} & \textbf{Best} & \(K\) & \(B\) & & 
\(\mathrm{P}\) & \textbf{Avg} & \textbf{Best} & \(K\) & \(B\) & &
\(\mathrm{P}\) & \textbf{Avg} & \textbf{Best} & \(K\) & \(B\) \\
\midrule
400  & 46005 & 46005 & 10 & 7  & & 400  & 60730 & 60730 & 13 & 10 & & 400  & 73240 & 73240 & 16 & 12 \\
600  & 43219 & 43219 & 10 & 6  & & 600  & 57921 & 57921 & 13 & 9  & & 600  & 72057 & 72057 & 15 & 12 \\
800  & 40248 & 40248 & 10 & 5  & & 800  & 50078 & 50078 & 14 & 6  & & 800  & 66915 & 66915 & 16 & 10 \\
1000 & 37985 & 37985 & 11 & 5  & & 1000 & 45143 & 45143 & 15 & 4  & & 1000 & 60546 & 60546 & 16 & 8  \\
1200 & 34080 & 34080 & 10 & 6  & & 1200 & 39527 & 39527 & 13 & 3  & & 1200 & 54849 & 54849 & 16 & 6  \\
1400 & 28656 & 28656 & 11 & 1  & & 1400 & 33537 & 33537 & 13 & 1  & & 1400 & 39373 & 39373 & 16 & 1  \\
1600 & 24372 & 24372 & 10 & 2  & & 1600 & 30252 & 30252 & 13 & 0  & & 1600 & 35238 & 35238 & 15 & 0  \\
\bottomrule
\end{tabular}
\end{table}

%% file: table/Exp5RealWorld.tex
\begin{table}[H]
\centering
\caption{Real-World Hospital Instances in Singapore}
\label{tab:real_world_data}
\begin{tabular}{lccccc}
\toprule
\textbf{Data Instance} & \textbf{Avg Cost} & \textbf{Best Cost} & \textbf{Best Time} & \textbf{Truck \#} & \textbf{Charger \#} \\
\midrule
10\_hospital & 4152 & 4152 & 183893  & 2 & 0 \\
11\_hospital & 4050 & 3939 & 222894  & 2 & 0 \\
18\_hospital & 7215 & 7215 & 5137671 & 2 & 1 \\
23\_hospital & 8466 & 8466 & 7352512 & 3 & 1 \\
26\_hospital & 8880 & 8873 & 4875869 & 3 & 1 \\
29\_hospital & 9103 & 9097 & 6330522 & 3 & 1 \\
\bottomrule
\end{tabular}
\end{table}

%% file: Sections/7Conclusion.tex
\section{Conclusion}

In this paper, we proposed a WMC-EVRP model, focusing on reducing the number of MTEV and MCT usage while targeting the shortest routes. Our model opens up a new, more efficient, and flexible field, the on-road charging method for EVRP, instead of the existing charging stations and battery swapping. 

We developed a mathematical model that combines the representation of the behavior characteristics of the traverse of both MTEV and MCT. This model takes into account varying speeds and energy consumption rates, providing a more realistic and practical framework for optimizing routes. To solve this complex synergistic problem, we proposed the LNS-LS-BDP algorithm, which identifies the optimal routes for all vehicles, ensuring that both delivery and charging operations are carried out in the most efficient manner while meeting all demands. Our experimental analysis included test instances with more than 300 customers, and we also validated our approach using real-world data from Singapore hospitals, ensuring the practical applicability of our model and algorithm in real logistics scenarios. The results of these experiments show that our Synergized Mobile Charging strategy not only reduces the number of vehicles needed, but also significantly optimizes energy usage, contributing to more sustainable and cost-effective transportation solutions.

The implications of our research extend beyond the immediate contributions to the field of EVRP. By focusing on the integration of mobile charging solutions with route optimization, we offer a framework that are suited to the emerging future trends of EV technology. As the adoption of EVs continues to accelerate, the demand for innovative and efficient charging solutions will become increasingly critical. Our work addresses this need by providing a scalable and adaptable approach that can be applied to broader logistics and transportation networks other than medical deliveries. 

Looking ahead, our WMC-EVRP framework opens up numerous opportunities for future research. One promising direction is the further refinement of our model to incorporate additional, widely-research constraints and objectives in traditional EVRP, such as time windows, stochastic demand, and dynamic traffic conditions. Additionally, the integration of real-time data and predictive analytics could enhance the responsiveness and adaptability of our algorithm, making it even more effective in dynamic and uncertain environments. Another important area for future exploration is the broader application of our framework to other types of EVs and industries. For example, the principles underlying our on-road mobile charging strategy could be adapted to optimize the operations of electric commercial buses, delivery drones, or even autonomous electric fleets. By extending our approach to these new contexts, researchers can continue to push the boundaries of what is possible in the field of EV routing and logistics. Moreover, as mobile charging technologies, such as wireless and on-road charging, become more prevalent, our framework is well-positioned to integrate these advancements. The ability to charge vehicles through powered highways while they are in motion will further enhance the flexibility and efficiency of our model, making it a valuable tool for future transportation systems. This integration could lead to new hybrid models that combine mobile charging with stationary infrastructure, providing solutions for the charging needs of mobile chargers, charging robots, and more EV derivatives.

%% file: Sections/8Appendice.tex
\appendix
\section*{Appendix}
\label{Appendix}

\input{table/Exp2RemovalAvgTime}

\input{table/Exp2InsertionAvgTime}

\input{table/Exp2RemovalAvgUpdate}

%% file: table/Exp2RemovalAvgTime.tex
\begin{table}[H]
\centering
\caption{Averaged Removal Running Times Across Different Data Sizes}
\label{tab:avg_removal_run_times}
\begin{tabular}{lcccccc}
\toprule
\textbf{Data Size} & \textbf{RR Time} & \textbf{DR Time} & \textbf{SR Time} & \textbf{WR Time} & \textbf{ShR Time} & \textbf{CR Time} \\
\midrule
50Avg  & 996.8  & 3149.5  & 1677.0  & 3077.1  & 13692.2 & 2745.8  \\
60Avg  & 1119.6 & 3540.6  & 2005.8  & 3612.9  & 18112.0 & 4185.9  \\
70Avg  & 1330.4 & 4189.7  & 2426.7  & 4359.1  & 27781.8 & 4680.0  \\
80Avg  & 1453.9 & 4995.8  & 3053.8  & 5293.5  & 32223.1 & 4976.0  \\
90Avg  & 1498.0 & 5412.2  & 2979.3  & 5213.9  & 32598.2 & 4595.1  \\
100Avg & 1467.8 & 5365.6  & 2984.5  & 5329.4  & 31071.5 & 4237.4  \\
110Avg & 1469.1 & 5733.2  & 3621.1  & 5788.8  & 33978.7 & 3802.5  \\
120Avg & 1709.8 & 6393.4  & 4134.9  & 6415.3  & 42229.6 & 4691.3  \\
130Avg & 1626.4 & 6405.7  & 4048.9  & 6102.6  & 42555.8 & 4409.1  \\
140Avg & 1941.1 & 7308.2  & 5067.0  & 7575.0  & 51390.2 & 4822.3  \\
150Avg & 2034.1 & 8130.0  & 6133.8  & 8686.7  & 68619.7 & 5676.5  \\
160Avg & 2463.7 & 10183.7 & 7230.5  & 10153.4 & 71252.4 & 5629.0  \\
170Avg & 2606.1 & 12568.9 & 9158.1  & 12085.7 & 100091.0 & 6022.3 \\
\bottomrule
\end{tabular}
\end{table}

%% file: table/Exp2InsertionAvgTime.tex
\begin{table}[H]
\centering
\caption{Averaged Insertion Running Times Across Different Data Sizes}
\label{tab:avg_iteration_times}
\begin{tabular}{lcccccc}
\toprule
\textbf{Data Size} & \textbf{RI Time} & \textbf{GI Time} & \textbf{SI Time} & \textbf{R2I Time} & \textbf{R3I Time} & \textbf{CI Time} \\
\midrule
50Avg  & 1701.0  & 7804.1  & 2584.3  & 1217641.7  & 1158857.6  & 23122.4  \\
60Avg  & 1923.3  & 10226.8 & 2977.2  & 2021280.3  & 1874620.8  & 28131.4  \\
70Avg  & 2055.4  & 14069.0 & 3665.2  & 2753740.8  & 2680727.6  & 31348.0  \\
80Avg  & 2536.5  & 17821.2 & 4534.5  & 3786755.3  & 4040915.8  & 38374.5  \\
90Avg  & 2385.9  & 19361.0 & 4959.5  & 5303871.1  & 4869916.7  & 41047.4  \\
100Avg & 2013.7  & 16480.9 & 4148.0  & 5253019.7  & 5163282.9  & 39591.6  \\
110Avg & 2088.4  & 19789.3 & 4330.8  & 6772482.4  & 7092118.0  & 41060.0  \\
120Avg & 2149.1  & 27706.9 & 5779.6  & 8629167.7  & 8085857.0  & 55346.8  \\
130Avg & 2038.5  & 24050.7 & 5042.4  & 9374281.4  & 9555169.1  & 47472.9  \\
140Avg & 2339.2  & 29899.5 & 6136.0  & 12127510.7 & 11464018.7 & 55558.6  \\
150Avg & 2783.7  & 33944.1 & 7092.6  & 16902284.6 & 16477536.4 & 68852.5  \\
160Avg & 3158.6  & 43178.4 & 8620.5  & 20716852.1 & 20449004.8 & 76290.5  \\
170Avg & 3558.0  & 64911.8 & 12339.3 & 28977484.5 & 27533500.6 & 99300.5  \\
\bottomrule
\end{tabular}
\end{table}

%% file: table/Exp2RemovalAvgUpdate.tex
\begin{table}[H]
\centering
\caption{Averaged Updates Per 10,000 Usage Across Different Data Sizes}
\label{tab:avg_updates_combined}
\begin{tabular}{lcccccccccccc}
\toprule
\textbf{Data Size} & \textbf{RR} & \textbf{DR} & \textbf{SR} & \textbf{WR} & \textbf{ShR} & \textbf{CR} & \textbf{RI} & \textbf{GI} & \textbf{SI} & \textbf{R2I} & \textbf{R3I} & \textbf{CI} \\
\midrule
50Avg  & 0.08  & 0.43  & 0.27  & 1.12  & 5.83  & 1.37  & 0.13  & 1.07  & 0.16  & 10.79  & 10.61  & 0.09  \\
60Avg  & 0.12  & 0.51  & 0.34  & 1.29  & 6.31  & 1.57  & 0.19  & 1.32  & 0.20  & 7.61  & 7.40  & 0.13  \\
70Avg  & 0.17  & 0.59  & 0.43  & 1.49  & 6.87  & 1.80  & 0.28  & 1.58  & 0.24  & 6.03  & 5.81  & 0.17  \\
80Avg  & 0.23  & 0.68  & 0.53  & 1.72  & 7.52  & 2.05  & 0.38  & 1.84  & 0.29  & 5.07  & 4.82  & 0.21  \\
90Avg  & 0.29  & 0.78  & 0.63  & 1.96  & 8.23  & 2.33  & 0.49  & 2.10  & 0.34  & 4.49  & 4.20  & 0.25  \\
100Avg & 0.37  & 0.88  & 0.74  & 2.22  & 9.00  & 2.63  & 0.63  & 2.36  & 0.40  & 4.14  & 3.82  & 0.30  \\
110Avg & 0.46  & 0.98  & 0.86  & 2.49  & 9.83  & 2.96  & 0.78  & 2.63  & 0.47  & 3.91  & 3.55  & 0.35  \\
120Avg & 0.56  & 1.09  & 0.99  & 2.78  & 10.71  & 3.32  & 0.95  & 2.89  & 0.53  & 3.77  & 3.38  & 0.40  \\
130Avg & 0.67  & 1.20  & 1.13  & 3.09  & 11.64  & 3.71  & 1.14  & 3.16  & 0.61  & 3.66  & 3.25  & 0.46  \\
140Avg & 0.80  & 1.32  & 1.28  & 3.42  & 12.62  & 4.12  & 1.34  & 3.43  & 0.69  & 3.58  & 3.16  & 0.52  \\
150Avg & 0.93  & 1.44  & 1.44  & 3.77  & 13.65  & 4.56  & 1.57  & 3.69  & 0.78  & 3.52  & 3.09  & 0.58  \\
160Avg & 1.08  & 1.56  & 1.61  & 4.13  & 14.73  & 5.02  & 1.81  & 3.96  & 0.87  & 3.47  & 3.03  & 0.65  \\
170Avg & 1.24  & 1.69  & 1.79  & 4.51  & 15.86  & 5.50  & 2.07  & 4.23  & 0.96  & 3.43  & 2.99  & 0.73  \\
\bottomrule
\end{tabular}
\end{table}